\UseRawInputEncoding
\documentclass{article}
\usepackage{float}
\usepackage[margin=1.0in]{geometry}
\usepackage{abstract}
\usepackage{afterpage}
\usepackage[backend=bibtex,style=nature,natbib=true,maxbibnames=99,minalphanames=3]{biblatex}
\usepackage{hyperref}
\newcommand{\HandE}{H\texorpdfstring{\&}{&}E}
\usepackage{caption}
\usepackage{lmodern}
\usepackage{threeparttable}
\usepackage[dvipsnames,svgnames, table,xcdraw]{xcolor}
\usepackage{relsize}
\usepackage{tcolorbox}
\usepackage{listings}
\usepackage{subcaption}
\usepackage{algorithm}
\usepackage{algorithmic}
\usepackage{comment}
\usepackage{graphicx}
\usepackage{booktabs}
\usepackage{multirow}
\usepackage{todonotes}
\usepackage{enumitem}
\usepackage{url}
\usepackage{mathpazo}
\usepackage{amsmath,amssymb,amsthm,amsfonts,bm}
\usepackage[toc,page,header]{appendix}
\usepackage{fancyhdr}
\usepackage[T1]{fontenc}

\usepackage{auth_detailed}

\counterwithout{figure}{section}
\counterwithout{table}{section}
\definecolor{darkgoldenrod}{rgb}{0.72, 0.53, 0.04}
\definecolor{backgroundcolor}{RGB}{250, 250, 252}   
\definecolor{keywordcolor}{RGB}{30, 0, 178}       
\definecolor{stringcolor}{RGB}{204, 0, 102}        
\definecolor{numbercolor}{RGB}{0, 128, 128}        
\definecolor{emphcolor}{RGB}{30, 0, 178}            
\definecolor{commentcolor}{RGB}{0, 128, 0}       
\definecolor{basiccodecolor}{RGB}{61, 61, 61}       

\lstdefinestyle{customstyle}{
    backgroundcolor=\color{backgroundcolor},   
    commentstyle=\color{commentcolor},
    keywordstyle=\color{keywordcolor},
    numberstyle=\color{numbercolor},
    stringstyle=\color{stringcolor},
    basicstyle=\color{basiccodecolor}\ttfamily\footnotesize,
    breakatwhitespace=false,         
    breaklines=true,                 
    captionpos=b,                    
    keepspaces=true,                 
    numbers=left,     
    basicstyle=\color{basiccodecolor}\ttfamily\footnotesize,
    numbersep=5pt,             
    xleftmargin=2em,
    xrightmargin=2em,
    showspaces=false,                
    showstringspaces=false,
    showtabs=false,                  
    tabsize=1,
    frame=single,
    framesep=5pt,
    framexleftmargin=1.5em,
    framexrightmargin=1.5em,
    framextopmargin=1pt,
    framexbottommargin=1pt,
    aboveskip=10pt,
    belowskip=10pt,
    breaklines=true,
    breakautoindent=true,
    emph={textgrad, tg, Variable, MultipleChoiceTestTime,
    TextualGradientDescent, BlackboxLLM},             
    emphstyle={\color{emphcolor}},
    extendedchars=true,
}

\usepackage[normalem]{ulem}
\usepackage{xcolor}

\newcommand{\scrcmd}{\textsuperscript}
\newcommand{\scr}[1]{\scrcmd{#1}}

\definecolor{logocolor}{RGB}{30, 0, 178}                

\newcommand{\COMMENTT}[1]{\textcolor{ForestGreen}{\# \texttt{#1}}}

\definecolor{darkerlogocolor}{RGB}{20, 0, 145}  

\newtcolorbox{ttcolorbox}[1][]{colframe=darkerlogocolor, colback=darkerlogocolor!4!white, title=#1}

\newtcolorbox{apxtcolorbox}[1][]{colframe=black, colback=black!3!white, title=#1}

\usepackage{fontawesome5}
\usepackage{helvet}
\DeclareRobustCommand{\corrAuthor}{\textsuperscript{\faEnvelope[regular]}}

\hypersetup{
  colorlinks=true,
  urlcolor=blue,
  citecolor=ForestGreen,
  linkcolor=blue
}
\definecolor{tissuelabblue}{HTML}{6352a3}

\newcommand{\arialtitle}[1]{{\fontfamily{phv}\selectfont #1}}

\title{{\fontsize{15pt}{18pt}\selectfont \textbf{\arialtitle{A co-evolving agentic AI system for medical imaging analysis}}}}

\author{\name Songhao Li$^{1,2}$,
\name Jonathan Xu$^{3,*}$,
\name Tiancheng Bao$^{2,*}$,
\name Yuxuan Liu$^{4,*}$,
\name Yuchen Liu$^{2,*}$,
\name Yihang Liu$^{2,*}$,
\name Lilin Wang$^{2}$,
\name Wenhui Lei$^{1}$,
\name Sheng Wang$^{1}$,
\name Yinuo Xu$^{5}$,
\name Yan Cui$^{1,4}$,
\name Jialu Yao$^{1,2}$,
\name Shunsuke Koga$^{1}$,
\name Zhi Huang$^{1,6,}$\corrAuthor\\\newline
$^{1}$ Department of Pathology and Laboratory Medicine, University of Pennsylvania\\
$^{2}$ Department of Electrical and System Engineering, University of Pennsylvania\\
$^{3}$ The Wharton School, University of Pennsylvania\\
$^{4}$ Department of Bioengineering, University of Pennsylvania\\
$^{5}$ Department of Computer and Information Science, University of Pennsylvania\\
$^{6}$ Department of Biostatistics,
Epidemiology \texorpdfstring{\&}{&} Informatics, University of Pennsylvania\\\newline
*~~~Equal contribution\\
\corrAuthor~Correspondence: Zhi Huang (\href{zhi.huang@pennmedicine.upenn.edu}{\textcolor{tissuelabblue}{zhi.huang@pennmedicine.upenn.edu}}) \\
\begin{normalsize}
\begin{center} \href{https://github.com/zhihuanglab/TissueLab}{\raisebox{-0.1\height}{\includegraphics[height=1em]
{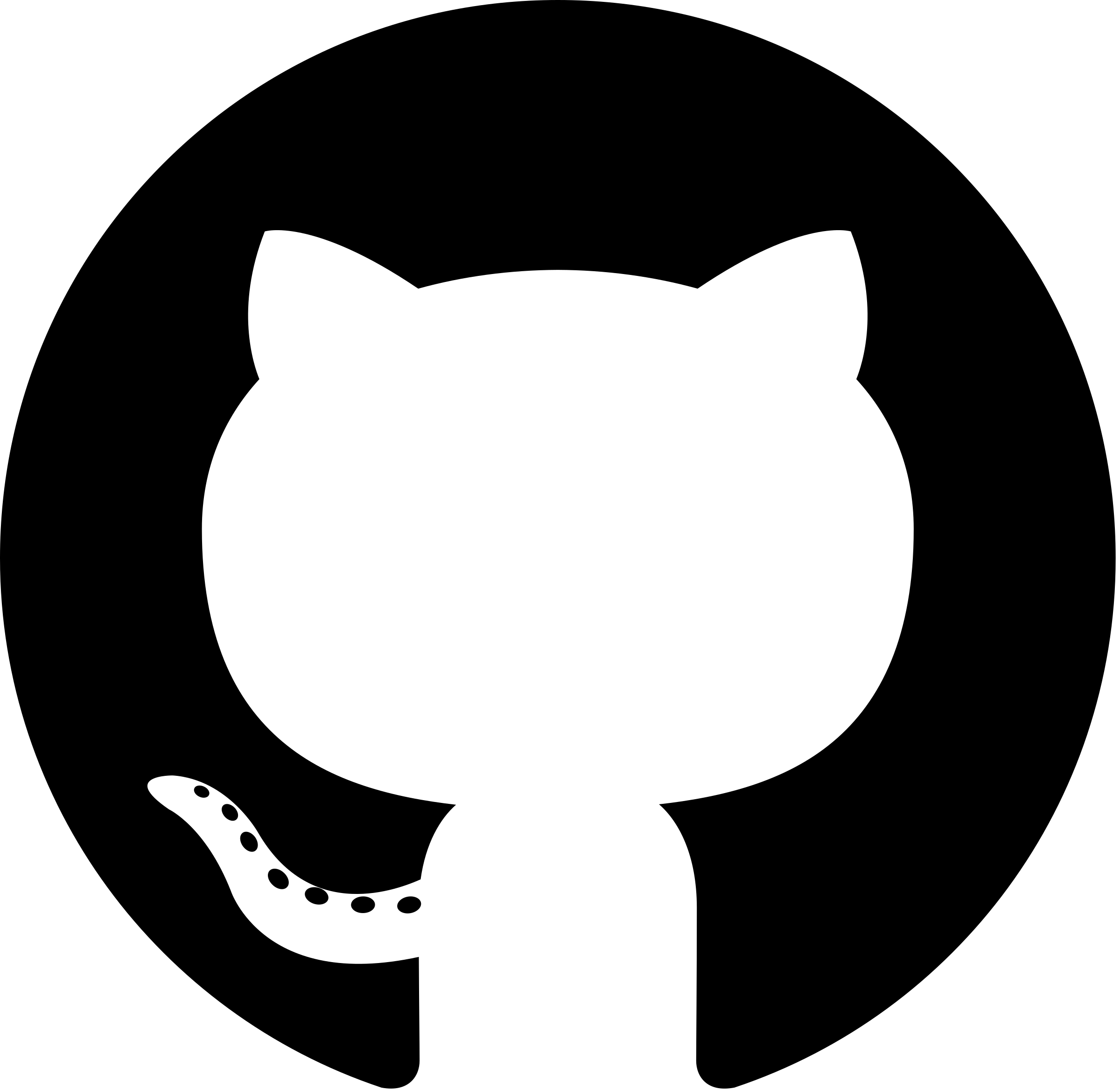}}
\textcolor{tissuelabblue}{Open-source code repository}} \hspace{20pt}
\href{https://www.tissuelab.org?from=arxiv}{\raisebox{-0.1\height}{\includegraphics[height=1em]
{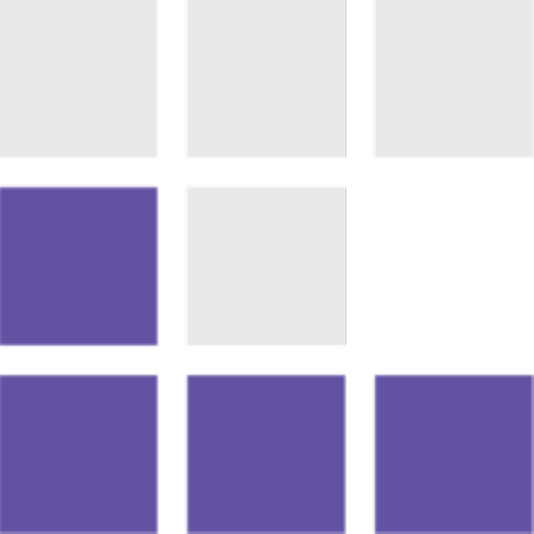}}
\textcolor{tissuelabblue}{TissueLab platform access}}
\end{center}
\end{normalsize}
}

\begin{document}

\fancyhead[L]{\raisebox{-0.1\height}{\includegraphics[height=0.9em]
{figures/TissueLab_logo.pdf} \href{https://www.tissuelab.org?from=arxiv}{\textsf{\textcolor{tissuelabblue}{tissuelab.org}}} }}
\fancyhead[R]{A co-evolving agentic AI system for medical imaging analysis}
\setlength{\headheight}{13pt}
\pagestyle{fancy}

\newcounter{suppfigure}
\newcounter{supptable}
\makeatletter
\newcommand\suppfigurename{Supplementary Figure}
\newcommand\supptablename{Supplementary Table}
\newcommand\suppfigureautorefname{\suppfigurename}
\newcommand\supptableautorefname{\supptablename}
\let\oldappendix\appendix
\renewcommand\appendix{%
    \oldappendix
    \setcounter{figure}{0}%
    \setcounter{table}{0}%
    \renewcommand\figurename{\suppfigurename}%
    \renewcommand\tablename{\supptablename}%
}
\makeatother

\maketitle

\renewcommand{\abstractnamefont}{\normalfont\normalsize\bfseries}
\renewcommand{\abstracttextfont}{\normalfont\normalsize}

\renewcommand{\abstractname}{Abstract}
\begin{abstract}
\abstracttextfont
Agentic AI is rapidly advancing in healthcare and biomedical research. However, in medical image analysis, their performance and adoption remains limited due to the lack of a robust ecosystem, insufficient toolsets, and the lack of real-time interactive expert feedback.
Here we present ``TissueLab'', a co-evolving agentic AI system that allows human to ask direct research questions, automatically plan and generate explainable workflows, and conduct real-time analyses where experts can visualize intermediate results and refine them.
TissueLab's ecosystem integrates tool factories spanning pathology, radiology, and spatial omics domains. By standardizing the inputs, outputs, capabilities, and use cases of diverse tools, TissueLab determines when and how to invoke these expert tools to address research and clinical questions.
Through experiments across diverse tasks, where clinically meaningful quantifications directly inform staging, prognosis, and treatment planning, we show that TissueLab achieves state-of-the-art performance compared with end-to-end vision-language models (VLM) and other agentic AI systems such as GPT-5.
Moreover, TissueLab ecosystem continuously learns from clinicians, accumulating knowledge and evolving toward improved classifiers and more effective decision strategies. With transparent model refinement through active learning, it can deliver accurate results in previously unseen disease contexts within minutes without requiring massive datasets or prolonged retraining.
In colon cancer, TissueLab reached 94.9\% accuracy in neoplastic cell quantification within 10-30 minutes of feedback with real-time updates, and in prostate cancer it adapted in two minutes to achieve 99.8\% accuracy in tumor-to-duct ratio measurement, outperforming state-of-the-art VLM baselines.
Released as a sustainable open-source ecosystem (\href{https://www.tissuelab.org?from=arxiv}{\textcolor{tissuelabblue}{tissuelab.org}}), we expect TissueLab to significantly advance and accelerate computational research and translational adoption in medical imaging, while establishing a foundation for the next generation of medical AI.
\end{abstract}

\section{Introduction}
Medical image analysis provides the computational foundation for understanding disease status and progression. It is essential for clinical decision-making, treatment planning, and, most importantly, advancing scientific discovery.
Traditionally, the design and implementation of comprehensive medical image analysis pipelines required extensive theoretical and computational expertise. These pipelines were expensive to build and challenging to adapt to diverse scientific questions. Clinicians often propose important investigative questions, but relied on collaboration with computational scientists to establish workflows and carry out the research. Such question- and hypothesis-driven computational analyses, reliant on manually constructed pipelines and slow feedback cycles, were not scalable and could not accelerate discovery at the pace required for next-generation autonomous biomedical research.

Advances in medical foundation models have provided stronger baselines for many downstream tasks, yet cancers and other complex diseases present highly variable morphologies, imaging modalities, often requiring carefully customized workflows across diverse study objectives, which makes fully automated analysis across diverse medical imaging scenarios challenging. At present, no single model reliably addresses the full spectrum of clinical imaging needs. Clinical quantifications remain central to prognosis and treatment planning. For example, evaluating the tumor-to-normal cell ratio in histology images is a critical yet tedious task that cannot be performed manually at scale. Current vision-language models (VLMs) often fail on such questions: they are not transparent about the computational process and are prone to hallucinations, which degrades performance, undermines trust, and ultimately leads to algorithm aversion~\cite{ferber_development_2025}.
Building effective medical AI therefore requires more than off-the-shelf foundation models~\cite{truhn_large_2023, chen_evidence-based_2025}. These models must be fine-tuned or embedded within more sophisticated, adaptive pipelines that can directly answer research and clinical questions in a reliable, reproducible, and transparent manner. The central challenge is translating rapidly evolving AI capabilities into safe, trustworthy systems that adapt to unseen scenarios without manual workflow design or extensive retraining.

Recent breakthroughs in large language models (LLMs) driven agentic AI~\citep{anthropic2024claude, yang2024qwen2, openai_gpt-oss-120b_2025, noauthor_gpt-5_2025, yang2024sweagent} demonstrated the potential to generalize across diverse tasks. By enabling flexible orchestration of tools through LLMs, these agents promise a path toward more universal AI applications that could be directly usable in clinical settings~\cite{litjens2017survey, wang_spatialagent_2025, swanson_virtual_2025}. However, current VLM-based agentic AI still faces intrinsic limitations. Most agents depend on a fixed toolbox of pretrained models~\cite{li_mmedagent_2024} that quickly become obsolete as the field evolves, requiring substantial maintenance and redevelopment costs. Moreover, hallucinations~\cite{kalai_why_2025, das_trustworthy_2025}, token overload, and attention dilution make it difficult for VLMs to safely coordinate multiple models or process high-dimensional data such as gigapixel whole-slide images or volumetric CT scans~\cite{wu_vision-language_2025}. Lastly, clinicians and researchers are unable to intervene and refine intermediate tools dynamically, limiting adaptability, performance, and trust in the overall workflow~\cite{dietvorst_algorithm_nodate}. Moreover, expert knowledge and preferences are not retained for future reuse. As a result, current agentic systems often reduce to pipelines of proprietary tools, which are costly to develop, fragile to maintain, and poorly suited for unseen clinical cases where accurate quantification can directly alter prognosis and treatment strategies. These gaps perpetuate the divide between AI researchers and clinical practitioners, limiting  research and real-world translational adoption.

To address these challenges, we introduce TissueLab, a co-evolving agentic AI system designed for medical imaging analysis that continuously evolves with new tools and user feedback. TissueLab emphasizes four principles: (i) \textbf{adaptivity}, realized through a modular plugin architecture that allows any state-of-the-art model to be mounted as a task node, further enhanced by semantic function-calling, which enables the agent to interpret and operate on heterogeneous data stored in the memory layer regardless of format, and by parallel workflow execution, which leverages topological sorting to distribute inference across independent branches so that total runtime is effectively bounded by the longest critical path rather than the sum of all tasks; (ii) \textbf{co-evolution}, by incorporating clinician feedback into active learning loops and maintaining an editable memory layer that persistently records intermediate results as training examples. Feedback provided by clinicians is transformed into labeled data that supports lightweight fine-tuning of downstream modules, enabling the system to rapidly adapt to clinical needs. In this way, the editable memory layer not only ensures transparency and traceability, but also serves as a reservoir of supervision that drives continual system refinement through feedback-driven fine-tuning; (iii) \textbf{safety}, as every diagnosis is explicitly grounded in authoritative clinical standards rather than unconstrained model reasoning. To achieve this, TissueLab employs the Model Context Protocol (MCP) to dynamically retrieve external guidelines and criteria ensuring that outputs are always supported by traceable references. In addition, all intermediate states are persisted within the system and remain fully transparent and visualizable to users, so that every decision can be reviewed, verified, and reproduced. By coupling external evidence retrieval with persistent transparency, TissueLab minimizes hallucinations and guarantees that its recommendations align with evolving standards of care. and (iv) \textbf{community value}, by enabling both AI researchers and clinicians to contribute models, annotations, and workflows within a shared ecosystem, thereby enhancing collective capability while allowing users to tailor the system to their own needs. Together, these principles position TissueLab as a bridge between rapid advances in AI research and their safe translation into reproducible, guideline-aligned, and clinically meaningful applications that remain responsive to evolving standards of care.

We demonstrate the versatility of TissueLab across pathology, radiology, and spatial omics, where clinically meaningful quantification directly informs staging, prognosis, and treatment planning. Beyond immediate clinical applications, TissueLab contributes to biomedical research by providing expert-level, scalable assessments that accelerate discovery and support cancer research and other disease studies. Through relational analyses and task-oriented evaluations, TissueLab enables universal yet trustworthy AI agents that bridge the divide between research innovation and clinical adoption-laying the foundation for an open, sustainable ecosystem that evolves in tandem with both medical needs and technological progress.

TissueLab is released as an open-source software available across Windows, macOS, and Linux. The online ecosystem is also available at \href{https://www.tissuelab.org?from=arxiv}{\textcolor{tissuelabblue}{tissuelab.org}}. All clinicians and researchers are free to access to the system, perform medical imaging analysis and annotations, contribute AI models, and co-develop agentic AI workflows. This commitment to openness reflects our team's mission to advance open-source medical AI and community value, ensuring that medical AI evolves collaboratively, sustainably, and in alignment with day-to-day practice.

\section{Results}
\label{sec:benchmarks}
\begin{figure}[hbtp]
    \centering
    \includegraphics[width = \textwidth]{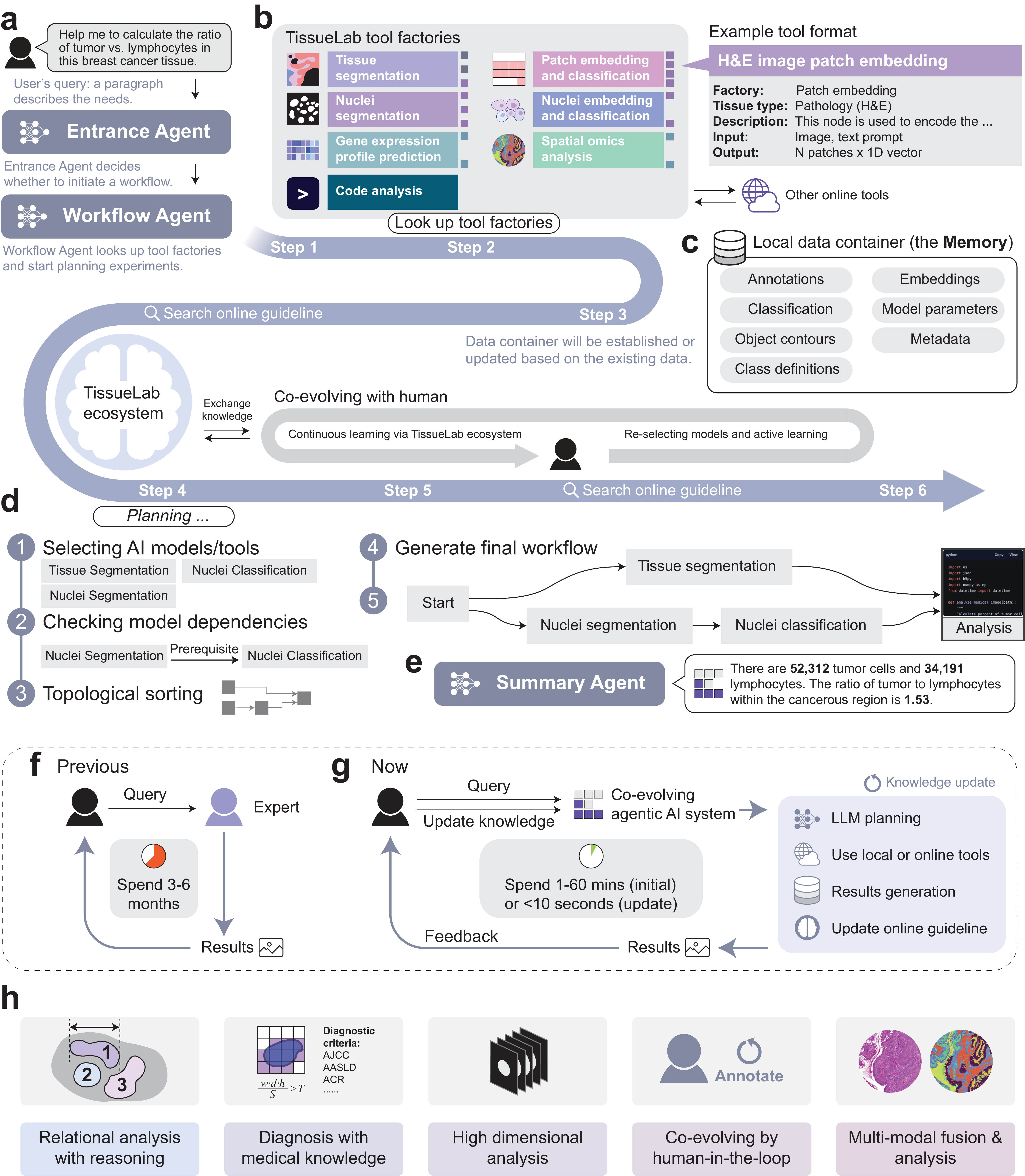}
    \caption{\small \textbf{Overview of the TissueLab agentic AI ecosystem for accelerating biomedical research and clinical decision support}. \textbf{a}. Orchestration phase in which the large language model plans workflows. \textbf{b}. Modular plugin architecture enabling seamless integration of new models without maintaining a fixed toolbox. \textbf{c}. Editable memory layer that supports co-evolution, allowing clinician feedback to align outputs with expert expectations and mitigate attention dilution. \textbf{d}. Execution phase where modular task nodes (e.g., segmentation, classification, spatial analysis) are scheduled via topological sorting, supporting both serial and parallel execution. \textbf{e}. Natural language summary agent for user-facing interpretation of results. \textbf{f}. Traditional clinical workflow, in which queries often required 3-6 months from initial question to expert-derived results. \textbf{g}. TissueLab workflow, where the same queries can be addressed within 1-60 minutes for initial setup and updated in less than 10 seconds after feedback. \textbf{h}. Representative applications, including relational analysis with reasoning, diagnosis with medical knowledge, high dimensional analysis, and multi-modal fusion and analysis.
}
    \label{fig:1}
\end{figure}

\subsection{Creating a co-evolving agentic AI system for medical imaging}
\label{sec:Creating a co-evolving agentic AI system for medical imaging}

We developed TissueLab, the first co-evolving agentic AI ecosystem for medical image analysis.
Inspired by the Factory Method design pattern~\cite{gof1994}, TissueLab abstracts diverse imaging problems into a small set of fundamental operations such as tissue segmentation, classification, and local code analysis, thereby enabling clinicians to access advanced AI analysis without requiring AI research or coding expertise.

\begin{figure}[hbtp]
    \centering
    \includegraphics[width = \textwidth]{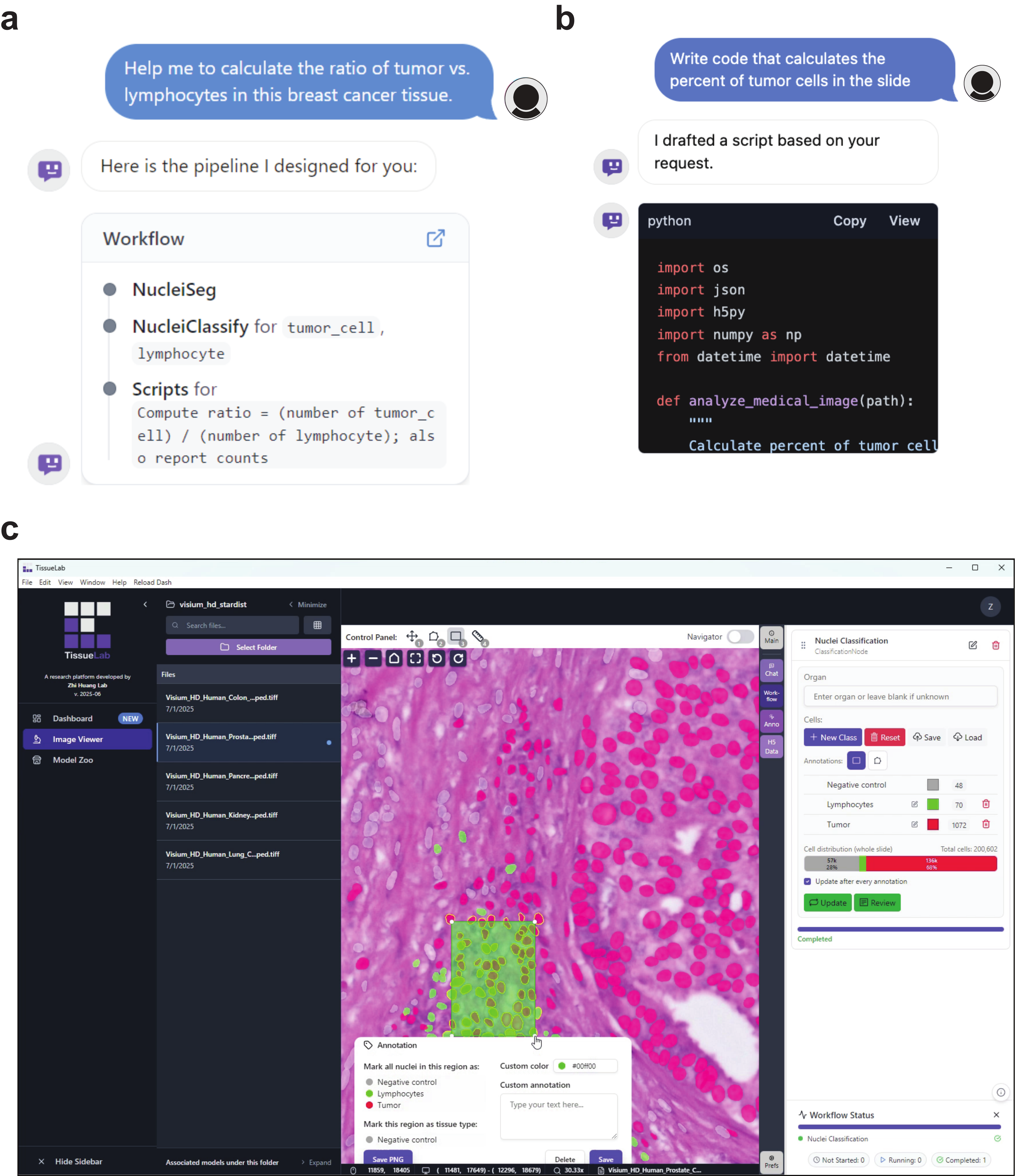}
    \caption{\small \textbf{TissueLab agentic AI system in use.} 
    \textbf{a}, Conversational interface of the agentic AI system, in which the agent produces a corresponding workflow that can be executed to obtain the cell proportion.
    \textbf{b}, Code-inspection interface for reviewing and editing generated code.
    \textbf{c}, User interface showing a whole-slide image (WSI) with cell segmentation results, where the user performs active learning-based classification to refine model predictions.}
    \label{supplementary:1}
\end{figure}

The TissueLab agentic AI system starts with users' query, e.g., ``\textit{Calculate the ratio of tumor cells vs. lymphocytes in this breast cancer tissue.}'' Based on the user query, our \textbf{entrance agent} will ask \textbf{workflow agent} to initiate a workflow (\textbf{Figure \ref{fig:1}a}). The available local tool factories, along with the capability of integrating new tools (\textbf{Figure \ref{fig:1}b}) without modifying the current codebase, as well as storing the intermediate results into a local data container (\textbf{Figure \ref{fig:1}c}), offers a practical solution for LLM as an orchestrator to invoke appropriate AI tools and avoid token overload\footnote{Token overload is avoided by storing all intermediate model results (e.g., NumPy arrays, CSV files) in an HDF5 file, allowing the analysis code to access and reuse them later.}. When selecting AI tools, our agentic AI system will check model denpendencies according to the model card, and perform topolocial sorting to enable distributed inference\footnote{Ditributed inference allows TissueLab agentic AI system to handle high-throughput requests while efficiently organizing tasks.} (\textbf{Figure \ref{fig:1}d}). Once the workflow has been generated (\textbf{Supplementary Figure \ref{supplementary:1}a}), TissueLab will directly analyze the results stored at the local data container, and the \textbf{summary agent} will produce a direct answer in natural language (\textbf{Figure \ref{fig:1}e}).

With TissueLab platform, user can further provide additional feedback through chat box therefore adjust/improve the analysis module (\textbf{Supplementary Figure \ref{supplementary:1}b}). During the workflow generation and execution in \textbf{Figure \ref{fig:1}d}, all intermediate results can be visualized and assessed by human experts through our software platform (\textbf{Supplementary Figure \ref{supplementary:1}c}). This interactive component is the key to enable the ``co-evloving'' function, which allows human experts to (i) provide expert annotations (\textbf{Supplementary Figure \ref{supplementary:1}c}); (ii) improve AI tools in real-time through iterative feedback (\textbf{Supplementary Figure \ref{supplementary:2}}); (iii) adjusting or reselecting appropriate models; and (iv) contribute new knowledge or reuse others knowledge in our TissueLab ecosystem.

Building on top of the fixed foundation models, TissueLab AI system enables users to contribute annotations to build lightweight classifiers that can be shared within the community and be reused immediately through active learning~\cite{huang_pathologistai_2025}. Compared with previous and traditional AI research workflow, where clinicians asks expert AI scientist to design and implement a computational pipeline, which is often a time-consuming process (\textbf{Figure \ref{fig:1}f}), TissueLab can now autonomously generate and execute expert-level workflows within 60 minutes (\textbf{Figure \ref{fig:1}g}). Moreover, the process can be iteratively refined with expert feedback in just 10 seconds per iteration. By incorporating human feedback, clinicians can actively participate in workflow design rather than remain passive users within the ecosystem, enabling the agentic AI system to continuously improve and become smarter over time.

Through comprehensive experimental evaluations in pathology, radiology, and spatial omics, we demonstrated the advancements of the TissueLab system, as detailed in later sections: (i) performing relational analysis with reasoning, (ii) enabling trustworthy diagnosis informed by medical knowledge, (iii) addressing high-dimensional data challenges, (iv) co-evolving with human experts, and (v) understanding and solving multi-modal analysis tasks (\textbf{Figure \ref{fig:1}h}).

\subsection{Co-evolving agentic AI accelerates expert-level assessment of tissue measurement}
\label{sec:Co-evolving agentic AI accelerates expert-level assessment of tissue measurement}
\begin{figure}[hbtp]
    \centering
    \includegraphics[width = \textwidth]{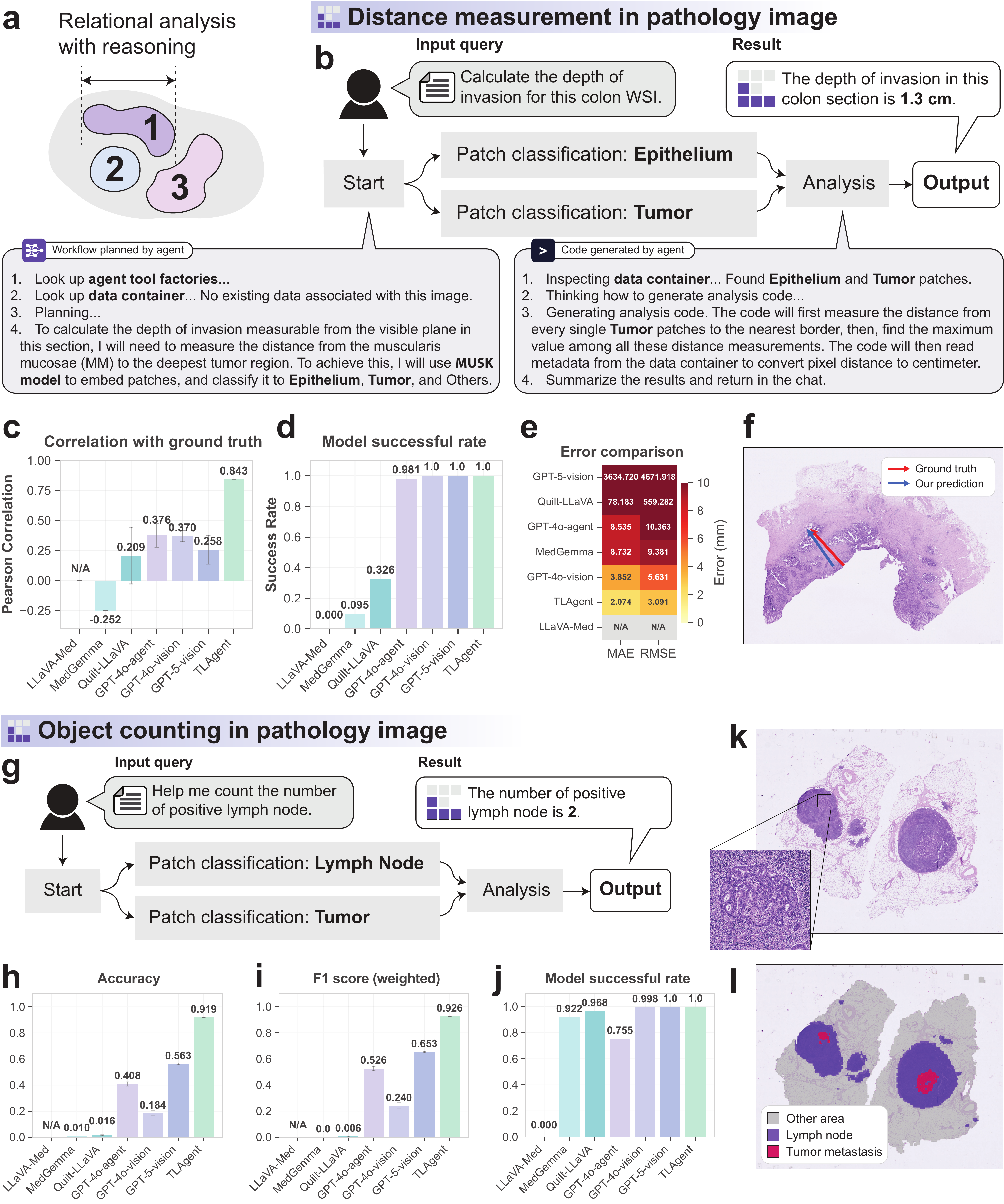}
    \caption{\small \textbf{Agentic AI performs relational analysis with reasoning}. \textbf{a}, Conceptual schematic of relational analysis with reasoning. \textbf{b}, Workflow generated by TLAgent for measuring depth of invasion (DoI) in colorectal cancer. \textbf{c}, Correlation with ground truth across methods for DoI. \textbf{d}, Task-completion (success) rate for DoI among baseline VLM agents versus TLAgent. \textbf{e}, Error metrics for DoI (MAE and RMSE, mm). \textbf{f}, An example of WSI showing expert annotation and TissueLab prediction overlaid, illustrating explainable alignment. \textbf{g}, Workflow generated by TLAgent for positive lymph-node counting in \HandE~pathology. \textbf{h}, Accuracy by method for lymph-node counting. \textbf{i}, Weighted F1-score (to account for class imbalance) for lymph-node counting. \textbf{j}, Task-completion (success) rate for lymph-node counting. \textbf{k}, Original WSI for the lymph-node task, serving as the ground truth reference. \textbf{l}, TissueLab segmentation overlaid on the whole slide images, providing explainable correspondence between predicted metastasis regions and expert labels.}
    \label{fig:2}
\end{figure}

\begin{figure}[hbtp]
    \centering
    \includegraphics[width = \textwidth]{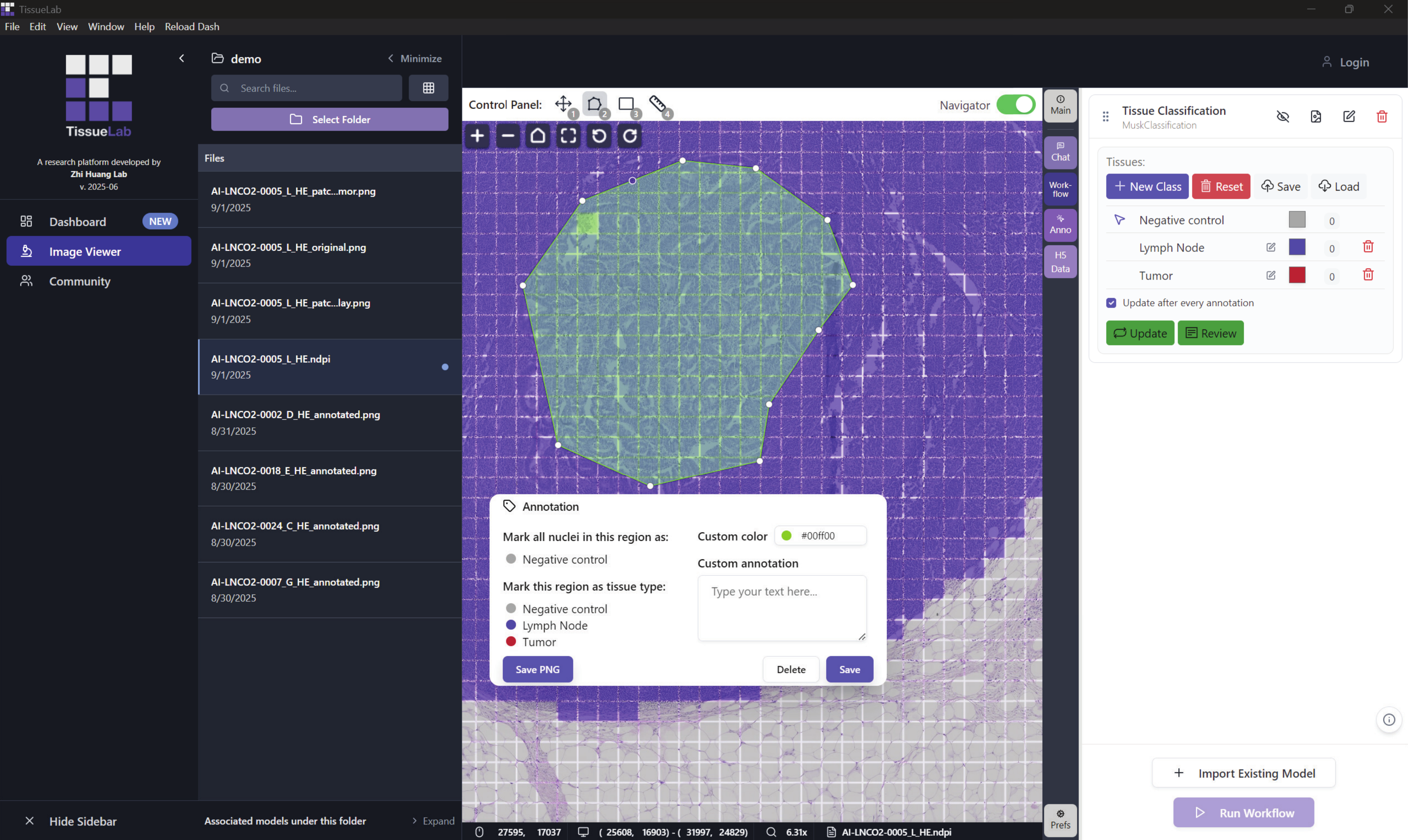}
    \caption{\small \textbf{ Tissue segmentation interface.} 
    User interface showing a whole-slide image (WSI) with tissue segmentation results, where the user performs active learning-based classification to refine model predictions.}
    \label{supplementary:2}
\end{figure}

Many clinically important measurements in pathology remain labor-intensive and lack suitable foundation models that can generalize across disease contexts. Tasks such as quantifying tumor invasion depth typically require expert pathologists to manually delineate tumor boundaries on whole-slide images, a process that is both time-consuming and difficult to scale (\textbf{Figure \ref{fig:2}a}). No current foundation model is specifically trained to address these fine-grained measurements. Moreover, training a dedicated vision-language model for each fine-grained measurement task would demand substantial data and computational resources, making it impractical in real-world clinical settings. Using TissueLab to automatically generates functional workflows and incorporates clinician annotations in a co-evolving loop (\textbf{Supplementary Figure~\ref{supplementary:2}}), lightweight modules can be rapidly updated and refined, sometimes within minutes, thereby ensuring that the system continuously aligns with clinical requirements. This adaptive strategy allows TissueLab to transform tasks traditionally dependent on extensive human labor into efficient, reproducible, and expert-level analyses.

We first evaluated TissueLab on predicting tumor invasion depth from whole-slide pathology images. Depth of invasion (DoI) is a critical histopathological feature that informs tumor staging and prognosis, and accurate and interpretable quantification is essential for clinical decision-making. Automatic DoI evaluation enables manual tasks to be completed autonomously and can support a wide range of clinical and research applications, such as Breslow thickness assessment in skin cancer. In this work, we study the DoI in regional lymph node metastasis in colon adenocarcinoma (LNCO2) dataset~\cite{maras_2020_lnco2} across 105 primary tumor resections from 49 patients.

As illustrated in \textbf{Figure~\ref{fig:2}b}, TissueLab agent (TLAgent) designs a structured workflow for measuring DoI in colorectal cancer. After the user query, the LLM orchestrator will look up agent tool factories, checking available data outputs, and come up with a workflow by invoking expert AI tools. After the process has finished, TLAgent will then perform automatic code analysis by inspecting data, generating and executing Python source code, and summarize the output in natural language.

As a result, the predicted invasion depths were strongly correlated with pathologist annotated ground truth (Pearson $\rho=0.843$), demonstrating robust consistency and superior performance across the full range of invasion severities compared to the second best model GPT-4o-agent (Pearson $\rho=0.376$) (\textbf{Figure~\ref{fig:2}c}). To ensure fair comparisons, we provided large thumbnail inputs for baseline models that could not process high-resolution whole-slide images and repeated each experiment five times. Nevertheless, some baseline models, such as Quilt-LLaVA, still failed under these conditions, either producing outputs unrelated to the query or explicitly returning no valid response. TLAgent consistently delivered the answer with 100\% in task completion (success) rate (\textbf{Figure~\ref{fig:2}d}). In terms of numerical accuracy, TLAgent achieved a mean absolute error (MAE) of 2.047~mm and a root mean square error (RMSE) of 3.091~mm relative to expert annotations, reflecting expert-level accuracy (\textbf{Figure~\ref{fig:2}e}).
While GPT-4o-vision reported competitive MAE values, its correlations with ground truth were poor (Pearson $\rho=0.37$), indicating that the predictions lacked directional consistency and often failed to reflect the true progression of invasion depth. For other approaches, MedGemma and GPT-4o-agent showed higher errors, with MAEs in the range of 8-9~mm; MedGemma also displayed a negative correlation with ground truth (Pearson $\rho=-0.22$). Both Quilt-LLaVA (MAE = 78.183 mm, RMSE $> 550$ mm) and GPT-5-vision (MAE = 3634 mm, RMSE = 4671 mm) exhibited excessively large errors, indicating that these predictions were highly unreliable, and LLaVA-Med often did not generate valid outputs. Note for all baseline models, experiments were repeated 5 times and the average performance was reported.

By taking a closer inspection on other approaches' result through TissueLab software, we found that these outputs were either random or lacking consistent pattern between predicted and true invasion depth. In contrast, TissueLab autonomously constructed a valid computational workflow: segmenting the relevant tumor tissue using domain-specific models, extracting the contours, and computing invasion depth as the maximum of the shortest distances between each tumor boundary point and the epithelial surface contour as detailed in Algorithm~\ref{appendix:doi-algorithm}~\cite{ajcc2024v9}. \textbf{Figure \ref{fig:2}f} shows that this approach reflects the clinical definition of the measurement and yields results that are not only numerically accurate but also directionally consistent with ground truth. These findings highlight the robustness of TissueLab’s co-evolving relational reasoning design in capturing clinically relevant spatial relationships.

\begin{algorithm}
\caption{Pseudocode for DOI measurement generated by TissueLab}
\label{appendix:doi-algorithm}
\begin{algorithmic}[1]
\STATE \textbf{Input:} Tumor boundary vertices $C_{\text{tumor}} = \{p_i\}_{i=1}^N$, Epithelial contour vertices $C_{\text{epi}} = \{e_j\}_{j=1}^M$, Pixel size $\text{pixel\_size}_{\mu m}$
\STATE \textbf{Initialize:} $D_{\max} = 0,\; p^\star = \text{None},\; q^\star = \text{None}$
\STATE \COMMENTT{Build spatial index (e.g., KD-tree) for $C_{\text{epi}}$}
\FOR{each $p \in C_{\text{tumor}}$}
    \STATE $q \gets \text{NearestPoint}(C_{\text{epi}}, p)$
    \STATE $d \gets \|p - q\|$ \COMMENTT{Euclidean distance in pixels}
    \IF{$d > D_{\max}$}
        \STATE $D_{\max} \gets d$
        \STATE $p^\star \gets p$
        \STATE $q^\star \gets q$
    \ENDIF
\ENDFOR
\STATE $\text{DOI}_{\mu m} \gets D_{\max} \times \text{pixel\_size}_{\mu m}$
\STATE \textbf{Output:} $\text{DOI}_{\mu m},\; p^\star,\; q^\star$
\STATE \COMMENTT{If epithelium has multiple contours, take $C_{\text{epi}}$ as the union of all vertices}
\end{algorithmic}
\end{algorithm}

We next evaluated the TissueLab agentic AI system on predicting the number of metastatic lymph nodes per slide in the regional lymph node metastasis in colon adenocarcinoma (LNCO2) dataset across 321 adjacent lymph node slides from 22 patients. This task is clinically significant for staging as the number of metastatic lymph nodes directly influences treatment decisions and the level of therapeutic aggressiveness ~\cite{nccn_2025_colon, gunderson_2010_jco, rosenberg_2008_annsur}. Since many slides do not contain metastatic lymph nodes, the dataset is imbalanced across categories. Therefore, the weighted F1-score is reported alongside overall accuracy.

As illustrated in \textbf{Figure~\ref{fig:2}g}, TLAgent accomplishes this task by first segmenting lymph nodes and tumor regions, then determining nodal positivity through code-based computation of spatial overlap between the two. This structured pipeline underlies the performance gains shown in \textbf{Figure~\ref{fig:2}h--i}. To be specific, TLAgent achieved accuracy = 0.919 and weighted F1 = 0.926, substantially outperforming baseline models Quilt-LLaVA (accuracy = 0.016, F1 = 0.006), MedGemma (accuracy = 0.010, F1 = 0.000), GPT-4o-vision (accuracy = 0.184, F1 = 0.240),  GPT-5-vision (accuracy = 0.563, F1 = 0.653), and GPT-4o-agent (accuracy = 0.408, F1 = 0.526). On another note, LLaVA-Med frequently failed to complete the task. Compared to the best-performing baseline (GPT-5-vision), TLAgent improved accuracy by +0.356 absolute points ($\approx 63\%$ relative gain) and weighted F1 by +0.273 ($\approx 41\%$ relative gain). In addition to its superior performance, TLAgent consistently achieved a 100\% task success rate (\textbf{Figure~\ref{fig:2}j}). Note for all baseline models, experiments were repeated 5 times and the average performance was reported.

These comparisons underscore a key distinction between TLAgent and existing agent-based or VLM-based models. Whereas single-model prompting often yields arbitrary or clinically meaningless outputs, TissueLab relies on co-evolving agentic orchestration to assemble domain-appropriate workflows. By integrating segmentation models, contour extraction, and geometric computation into a structured pipeline, the system ensures that its reasoning is clinically grounded and can be validated through visual inspection through the software. This design explains why TLAgent not only achieves lower error metrics but also produces results that align directionally with ground truth, highlighting the necessity of agentic orchestration for clinically meaningful relational reasoning. \textbf{Figure~\ref{fig:2}k--l} further illustrates both the original whole-slide image and intermediate results generated by TLAgent that can be visualized from TissueLab, showing that the system not only correctly counts positive lymph nodes but also identifies tissue types within each region and presents them through an interpretable visualization. Such explainable outputs position TissueLab as a clinically trustworthy assistant, fostering greater safety and confidence in the use of AI by pathologists.

\subsection{TissueLab enables guideline-aligned diagnosis using an effectively infinite medical knowledge database}

\subsubsection*{Leveraging online searching for diagnosis aligned with up-to-date clinical guidelines and expansive knowledge database}
Through the Model Context Protocol (MCP)~\cite{anthropic2024mcp}, TissueLab agents can access authoritative medical resources, such as AJCC~\cite{ajcc2024v9}, CAP~\cite{cap2023}, WHO~\cite{who2024haematolymphoid}, or AASLD~\cite{chalasani2018aasld} and other guidelines in real time. By incorporating these continuously updated knowledge sources and leveraging retrieval-augmented generation~\cite{lewis2020rag} in diagnostic reasoning, the system ensures that outputs remain aligned with current clinical standards. In contrast to models that rely on static internal knowledge, TissueLab dynamically incorporates external criteria, enabling guideline-consistent decisions for tasks such as metastasis classification or disease staging. In this way, the ecosystem effectively functions as an infinite and continuously updated knowledge base without any maintenance, while remaining anchored to authoritative clinical sources. This approach ensures that diagnostic outputs are not only accurate but also clinically safe, reproducible, and aligned with medical practice~\cite{topol_high-performance_2019}.

\subsubsection*{Assessment of lymph node metastasis using online diagnostic criteria}
\label{sec:Assessment of lymph node metastasis using online diagnostic criteria}
\begin{figure}[hbtp]
    \centering
    \includegraphics[width = \textwidth]{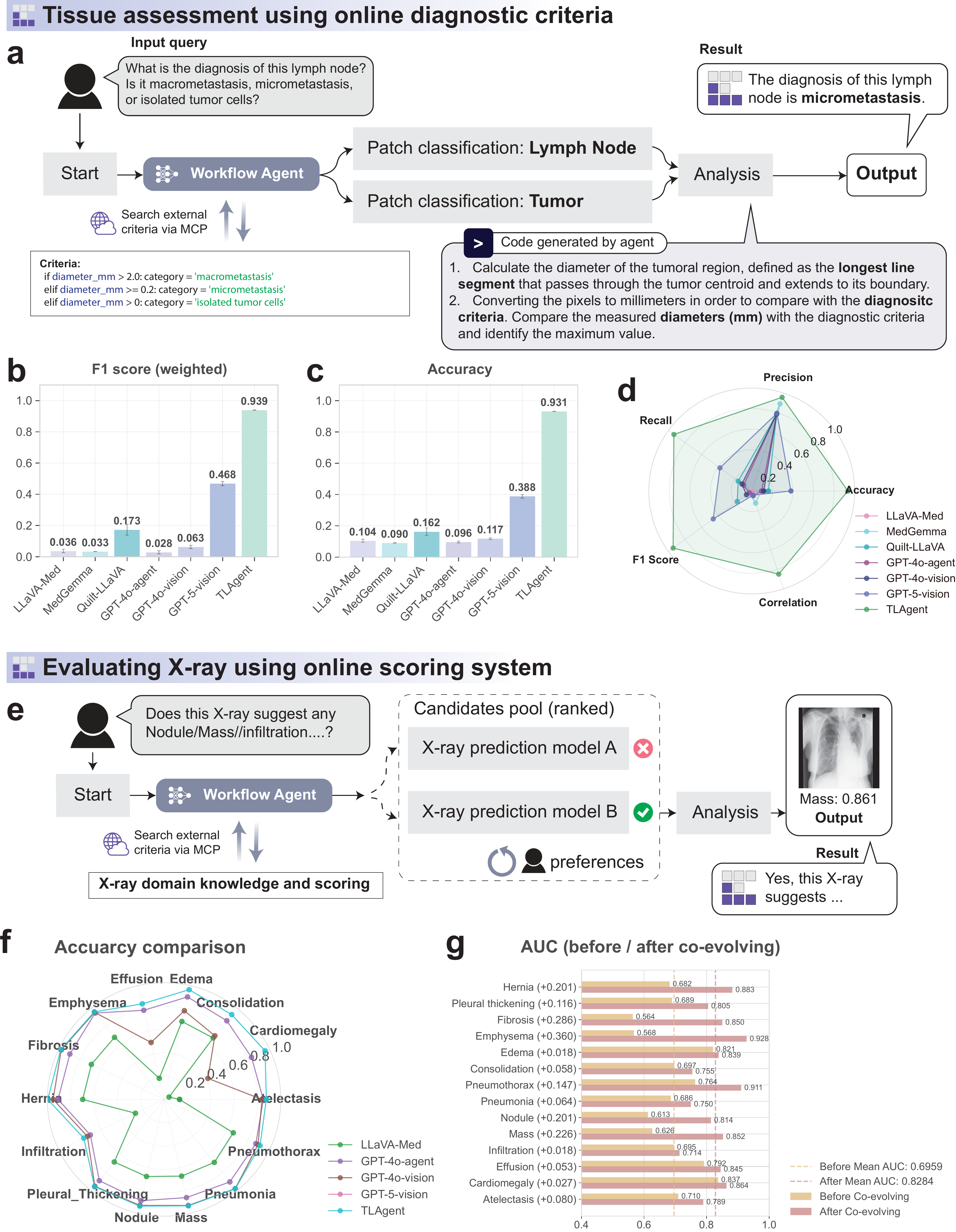}
    \caption{\small \textbf{Guideline-aligned diagnosis using agentic AI}. 
    \textbf{a}, Workflow for lymph-node metastasis classification in pathology. 
    \textbf{b}, Weighted F1-score comparison across models. 
    \textbf{c}, Accuracy comparison across models.
    \textbf{d}, Radar plot summarizing recall, precision, F1-score, accuracy, and correlation for metastasis classification. 
    \textbf{e}, Workflow for chest X-ray diagnosis by clinicians-in-the-loop co-evolution. 
    \textbf{f}, Per-disease accuracy radar across 14 conditions for each model. 
    \textbf{g}, AUC before and after co-evolution with clinician feedback, showing systematic improvement.
}
    \label{fig:3}
\end{figure}

To evaluate this capability, we further evaluated lymph node metastasis classification on the LNCO2 dataset across 321 lymph node slides, a staging-critical task that requires distinguishing macrometastasis ($> 2.0$ mm), micrometastasis ($\geq 0.2$ mm), and isolated tumor cells ($< 0.2$ mm)~\cite{ajcc2024v9}. These categories determine whether a lymph node is considered positive and thereby directly influence N stage assignment in the AJCC system and are clinically decisive for prognosis and the use of adjuvant therapy in colorectal and other cancers. For consistency across methods, baseline models that could not handle full whole-slide inputs were evaluated using large thumbnail representations, and each experiment was repeated five times. As illustrated in \textbf{Figure~\ref{fig:3}a}, TLAgent retrieved the relevant thresholds directly from these latest authoritative guidelines and applied them consistently to case measurements. Based on such criteria, the system achieved a weighted F1-score of 0.939 and an accuracy of 0.931, closely matching expert-level diagnosis (\textbf{Figure~\ref{fig:3}b--c}). Finally, \textbf{Figure~\ref{fig:3}d} summarizes recall, precision, F1-score, accuracy, and correlation in a radar plot, highlighting that TissueLab agentic AI system outperforms all baselines. Complete numerical results are provided in \textbf{Table~\ref{tab:multi-class-metrics}}.

\begin{table}[H]
\caption{Comparison of metastasis classification performance across models. Best performed model results are highlighted with bold font.}
\centering
\begin{threeparttable}
\begin{tabular}{@{}lccccc@{}}
\toprule
\textbf{Method} & \textbf{Accuracy} & \textbf{F1-score} & \textbf{Precision} & \textbf{Recall} & \textbf{Correlation} \\
\midrule
\textbf{TLAgent (ours)} & \textbf{0.931 $\pm$ 0.000} & \textbf{0.939 $\pm$ 0.000} & \textbf{0.951 $\pm$ 0.000} & \textbf{0.931 $\pm$ 0.000} & \textbf{0.849 $\pm$ 0.000} \\
GPT-5-vision  & $0.388 \pm 0.012$ & $0.468 \pm 0.013$ & $0.802 \pm 0.021$ & $0.388 \pm 0.012$ & $0.070 \pm 0.018$ \\
Quilt-LLaVA   & $0.162 \pm 0.023$ & $0.173 \pm 0.036$ & $0.780 \pm 0.042$ & $0.162 \pm 0.023$ & $-0.013 \pm 0.012$ \\
GPT-4o-vision & $0.117 \pm 0.006$ & $0.063 \pm 0.011$ & $0.783 \pm 0.025$ & $0.117 \pm 0.006$ & $-0.001 \pm 0.001$ \\
LLaVA-Med     & $0.104 \pm 0.010$ & $0.036 \pm 0.011$ & $0.733 \pm 0.178$ & $0.104 \pm 0.010$ & $0.015 \pm 0.006$ \\
GPT-4o-agent  & $0.096 \pm 0.006$ & $0.028 \pm 0.010$ & $0.710 \pm 0.442$ & $0.096 \pm 0.006$ & $0.048 \pm 0.004$ \\
MedGemma      & $0.090 \pm 0.000$ & $0.033 \pm 0.000$ & $0.887 \pm 0.000$ & $0.090 \pm 0.000$ & $0.125 \pm 0.000$ \\
\bottomrule
\end{tabular}
\caption*{\footnotesize Note: We report Cohen's $\kappa$ as the correlation metric because it measures agreement on categorical labels while correcting for chance. Pearson targets continuous variables; MCC is valid but less commonly interpreted in multi-class contexts.}
\end{threeparttable}
\label{tab:multi-class-metrics}
\end{table}

In contrast, baseline VLMs lack access and interpretation to external, up-to-date criteria therefore performed poorly across all metrics. For example, GPT-5-vision reached only moderate performance (F1 = 0.468, Accuracy = 0.388), while other widely used VLMs such as Quilt-LLaVA (F1 = 0.173, Accuracy = 0.162), LLaVA-Med (F1 = 0.036, Accuracy = 0.104), MedGemma (F1 = 0.033, Accuracy = 0.090), GPT-4o-agent (F1 = 0.028, Accuracy = 0.096), and GPT-4o-vision (F1 = 0.063, Accuracy = 0.117). From a clinical standpoint, such misclassifications are critical. These results demonstrate that TissueLab, by leveraging online search restricted to authoritative medical guidelines, produces clinically valid, guideline-aligned diagnostic outputs, instead of unguided single-model output.

\subsubsection*{Assessment of chest X-ray diagnosis using online guideline and human feedback}

Besides pathology image, TLAgent can also perform accurate and reliable chest X-ray diagnosis through online guideline retrieval and human feedback. Chest X-ray diagnosis task is clinically important but faces unique challenges: latest vision-language medical AI models are rapidly evolving, often surpassing older ones, yet their performance varies substantially across datasets and disease categories. To handle this variability, TissueLab incorporates a co-evolving mechanism that integrates clinician-in-the-loop oversight. As illustrated in \textbf{Figure~\ref{fig:3}e}, the system first generates a candidate pool of models for each query and selects the one with the best score. When clinicians provide negative feedback, TLAgent reselects from the same pool and updates model scores, allowing future rankings to better reflect clinical needs. In this way, the system continuously improves its clinical intelligence while ensuring interpretability.

We evaluated TLAgent on chest X-ray classification using the NIH Chest X-ray Dataset~\cite{wang2017chestxray8}, which comprises 112,120 frontal-view radiographs from 30,805 unique patients. Disease labels were derived by text-mining the original radiology reports using natural language processing.

While the dataset suitable for our experiment, accuracy alone can be misleading given severe class imbalance. Many thoracic conditions have low prevalence, so baseline VLMs appear to achieve high accuracy simply by defaulting to the negative class, without capturing true disease signals. As shown in the per-disease accuracy summary (\textbf{Table~\ref{tab:cxr-disease-performance}}) and the accuracy radar plot (\textbf{Figure~\ref{fig:3}f}), this will inflate overall accuracy but fail to capture clinically meaningful predictions.

~\begin{table}[H]
\caption{Comparison of disease classification performance across different agentic AI systems. Given the large size of the evaluation dataset, each experiment was conducted once, and mean or standard deviation values were not reported. Best performed model results are highlighted with bold font.}
\centering
\begin{tabular}{@{}lccccc@{}}
\toprule
\textbf{Disease} & \textbf{LLaVA-Med} & \textbf{GPT-4o-agent} & \textbf{GPT-4o-vision} & \textbf{TLAgent (ours)} \\ \midrule
Atelectasis        & 0.133 & 0.850 & 0.839 & \textbf{0.876} \\
Cardiomegaly       & 0.047 & 0.831 & 0.416 & \textbf{0.960} \\
Consolidation      & 0.674 & 0.862 & 0.697 & \textbf{0.929} \\
Edema              & 0.685 & 0.901 & 0.779 & \textbf{0.964} \\
Effusion           & 0.186 & 0.783 & 0.500 & \textbf{0.840} \\
Emphysema          & 0.681 & 0.947 & 0.954 & \textbf{0.959} \\
Fibrosis           & 0.691 & 0.896 & 0.978 & \textbf{0.983} \\
Hernia             & 0.697 & 0.909 & 0.951 & \textbf{0.997} \\
Infiltration       & 0.272 & 0.701 & 0.729 & \textbf{0.764} \\
Mass               & 0.674 & 0.862 & 0.932 & \textbf{0.934} \\
Nodule             & 0.677 & 0.889 & 0.931 & \textbf{0.937} \\
Pleural Thickening & 0.685 & 0.890 & 0.951 & \textbf{0.955} \\
Pneumonia          & 0.690 & 0.903 & 0.977 & \textbf{0.978} \\
Pneumothorax       & 0.658 & 0.872 & 0.893 & \textbf{0.912} \\
\bottomrule
\end{tabular}
\label{tab:cxr-disease-performance}
\end{table}

For this reason, the area under the ROC curve (AUC) provides a more reliable evaluation metric (\textbf{Figure~\ref{fig:3}g}). Unlike other VLM baselines which return categorical outputs without calibrated scores, our agentic AI framework generates step-wise prediction scores through adaptive orchestration (candidate pooling, best-score selection, and clinician-gated reselection\footnote{Effective clinician-gated reselection requires an online, open access ecosystem that can both retain user preferences and enable community-wide sharing. Achieving this was nontrivial, and at present this capability is uniquely implemented by TissueLab platform (\href{https://www.tissuelab.org?from=arxiv}{\textcolor{tissuelabblue}{\mbox{tissuelab.org}}}).} with preference updates). Such calibrated scores provide clinicians with interpretable confidence levels that can be directly referenced in decision-making, improving diagnostic safety and transparency. In turn, clinicians use these scores to provide feedback that guides the system's preference updates. This bidirectional loop establishes a genuine form of co-evolution: clinicians benefit from interpretable outputs, while their feedback continuously improves the agent's intelligence. With this design, TLAgent raised the mean AUC from 0.6959~\cite{tiu_expert-level_2022} to 0.8284~\cite{ma_fully_2025} across 14 thoracic conditions, leveraging a candidate pool mechanism to dynamically adapt and evolve with clinician's feedback. Other baseline models do not produce well-defined confidence scores, and therefore AUC cannot be computed.

\subsection{TissueLab enables high-dimensional radiology image analysis}
\label{sec:Agentic AI enables high-dimensional radiology image analysis}

\begin{figure}[tbp]
  \centering
  \includegraphics[width=\textwidth]{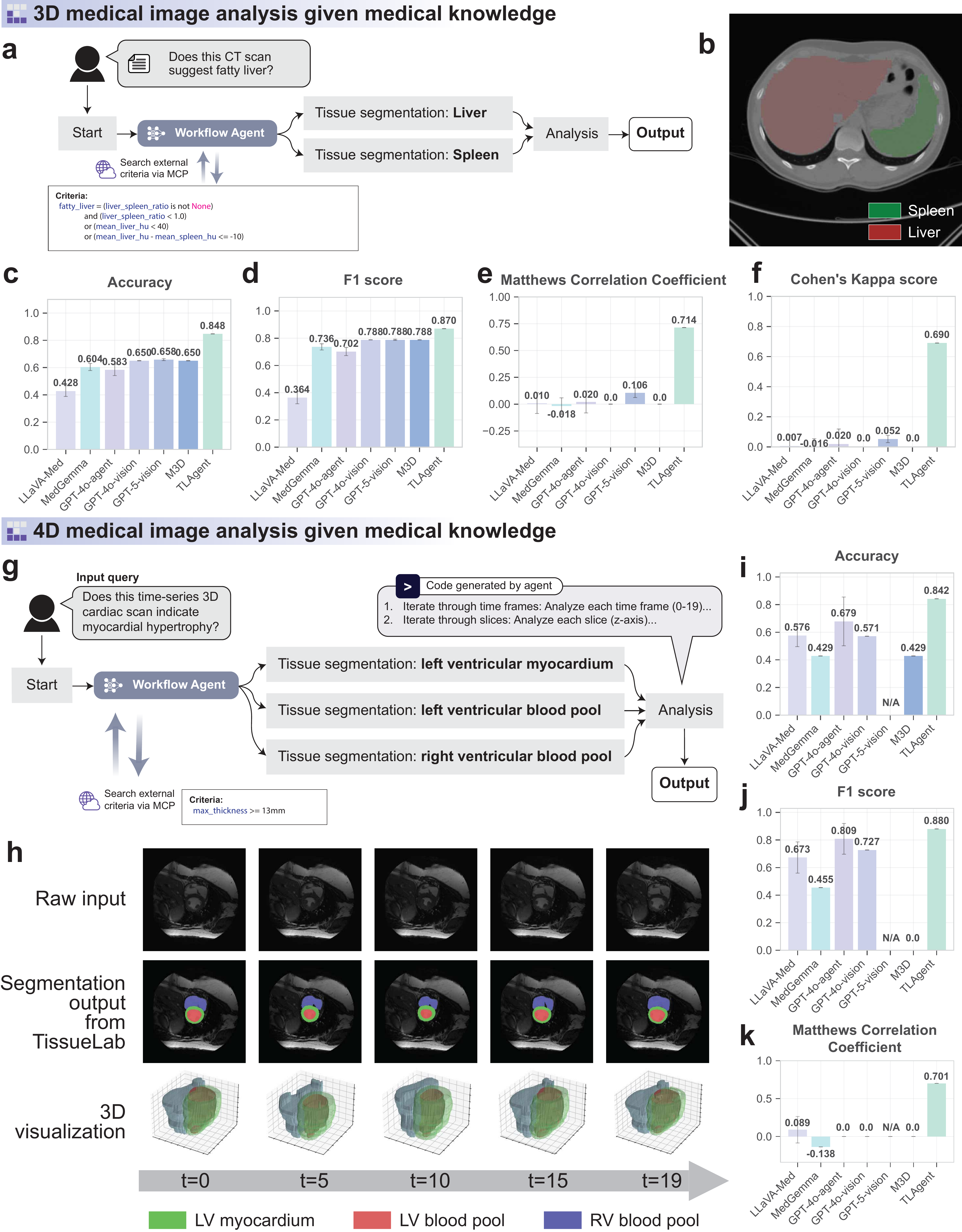}
\end{figure}

\afterpage{%
  \captionsetup{aboveskip=-10pt, belowskip=20pt}
  \captionof{figure}{\small \textbf{High-dimensional radiological analysis and diagnosis with medical knowledge.}
    \textbf{a}, Workflow diagram for fatty liver diagnosis. 
    \textbf{b}, Representative CT slice showing intermediate segmentation data during the TissueLab workflow. 
    \textbf{c}, Accuracy comparison across methods for fatty liver diagnosis. 
    \textbf{d}, Binary F1-score comparison for fatty liver diagnosis. 
    \textbf{e}, Matthews correlation coefficient for fatty liver diagnosis. 
    \textbf{f}, Cohen’s kappa score for fatty liver diagnosis. 
    \textbf{g}, Workflow diagram for myocardial hypertrophy assessment. 
    \textbf{h}, Representative sample showing intermediate segmentation outputs from time-series 3D cardiac image.
    \textbf{i}, Accuracy comparison of hypertrophy diagnosis. 
    \textbf{j}, Binary F1-score for hypertrophy diagnosis. 
    \textbf{k}, Correlation between predictions and ground truth.
}
  \label{fig:4}%
}

\subsubsection*{TLAgent predict fatty liver in 3D chest CT}

The TissueLab platform and TLAgent operate not only on 2D medical images but also on high-dimensional radiological data, such as volumetric and time-resolved imaging like CT and MRI scans.
One example is fatty liver in 3D chet CT. Fatty liver disease is a prevalent and clinically important condition associated with metabolic syndrome, cardiovascular risk, and long-term liver outcomes. In radiology practice, non-invasive diagnosis is based on guideline-defined Hounsfield unit (HU) values~\cite{chalasani2018aasld}. Here we evaluated TLAgent on the fatty liver diagnosis across 38 chest CT patients from held-out test set of the UNIFESP Chest CT Fatty Liver Competition Dataset~\cite{unifesp-fatty-liver}. As shown in \textbf{Figure~\ref{fig:4}a}, TLAgent autonomously retrieved the diagnostic criteria from AASLD and constructed a workflow that first segmented the liver and spleen volumes (\textbf{Figure~\ref{fig:4}b}). HU values were then computed from the segmented regions and evaluated against the guideline thresholds to determine the presence or absence of steatosis.

TLAgent demonstrated consistently high performance, with accuracy of 0.848 and F1-score of 0.870, supported by strong agreement-based metrics (MCC\footnote{MCC: Matthews correlation coefficient.} = 0.714, Cohen's kappa = 0.690) (\textbf{Figure~\ref{fig:4}c--f}). In contrast, most baseline models approached fatty liver diagnosis through either 2D slice-based recognition or simplistic code-based analysis, rather than leveraging true 3D medical imaging as TLAgent did, with the exception of the latest M3D agentic AI model. As a result, they showed weaker and often misleading performance. To ensure fair comparison, we manually provided 2D slices as inputs to these baselines and repeated the entire experiment five times. Accuracy values were 0.428 (LLaVA-Med), 0.604 (MedGemma), 0.583 (GPT-4o-agent), 0.650 (GPT-4o-vision), 0.658 (GPT-5-vision), and 0.650 (M3D). While some baselines produced comparable F1-scores (up to 0.788 for GPT-5-vision), their agreement-based metrics were nearly zero (\textbf{Figure~\ref{fig:4}e--f}). In particular, M3D and GPT-4o-vision defaulted to always predicting the positive class, which inflated recall and F1-scores but yielded no discriminative power (see \textbf{Supplementary Figure~\ref{supplementary:3}} for confusion matrices). These results highlight the advantage of agentic orchestration in the high-dimensional imaging tasks where these baselines may hallucinate plausible-seeming outputs that fail to align with clinical reliability.

\subsubsection*{TLAgent improves detection of intracranial hemorrhage from 3D CT scans}

In addition, TLAgent can also perform 3D brain MRI analysis. Here we demostrate the use case of identifying Intracranial hemorrhage (ICH). ICH is a life-critical neurological emergency in which rapid and accurate diagnosis is essential, as delays in detection can directly affect patient outcomes. However, reliable detection requires experts to examine hundreds of slices per scan, making the process time-consuming and prone to oversight. Automated systems capable of accurate and interpretable hemorrhage detection therefore have high clinical value. For benchmarking, we used the PhysioNet ICH dataset~\cite{hssayeni2020intracranial, hssayeni2020ct, goldberger2000physiobank}, which provides 82 3D CT scans with voxel-level annotations, enabling quantitative evaluation of hemorrhage segmentation and classification.

We ensured fair comparisons by evaluating baseline models on appropriate inputs and repeated each experiment five times. As illustrated in \textbf{Supplementary Figure~\ref{supplementary:4}a}, TLAgent searched authoritative diagnostic criteria for ICH and assembled a workflow that applied TotalSegmentator for volumetric hemorrhage segmentation and anatomical brain structure delineation, followed by code-based spatial overlap analysis to determine the presence of bleeding. This decomposition reflects the typical radiological reasoning process, beginning with hemorrhage localization and subsequently contextualizing the findings within neuroanatomical structures. Representative outputs of volumetric segmentation and slice-wise overlays are shown in \textbf{Supplementary Figure~\ref{supplementary:4}b} through TissueLab platform, providing transparent evidence for each diagnostic decision.

As quantified in \textbf{Supplementary Figure~\ref{supplementary:4}c--d}, TLAgent achieved an accuracy of 78.7\% with an F1-score of 0.750, closely matching expert-level performance. In contrast, general-purpose vision-language models struggled to provide clinically reliable outputs. LLaVA-Med effectively failed (F1 = 0.000) despite superficial accuracy of 52\%, while MedGemma (Acc = 45.3\%, F1 = 0.226) and GPT-4o-vision (Acc = 57.3\%, F1 = 0.273) produced unstable and anatomically implausible predictions across slices. GPT-4o-agent performed somewhat better (Acc = 51.0\%, F1 = 0.645) but still lagged behind TLAgent, reflecting frequent mislocalizations. GPT-5-vision and M3D were unable to complete the task. These comparisons emphasize that TLAgent was the only system able to combine high numerical accuracy with clinically coherent predictions.

For the more fine-grained sub-question of intraventricular hemorrhage detection in \textbf{Supplementary Figure~\ref{supplementary:4}e--f}, TLAgent maintained meaningful discriminative ability, achieving an F1-score of 0.500 and an accuracy of 0.947. However, LLaVA-Med and GPT-4o-vision both failed completely (F1 = 0.000), while MedGemma (F1 = 0.091, Acc = 0.733) and GPT-4o-agent (F1 = 0.084, Acc = 0.441) produced unstable or near-random predictions. GPT-5-vision and M3D were unable to provide evaluable outputs. Notably, although GPT-4o-vision appeared competitive in accuracy (0.896), its zero F1-score indicated predictions that were inconsistent and clinically meaningless.

\subsubsection*{TLAgent enables dynamic detection of myocardial hypertrophy from 4D cardiac imaging}
Lastly, in addition to 3D image, TLAgent can also handle 4D imaging data, where volumetric data are captured across sequential time points. For example, myocardial hypertrophy is a clinically critical condition associated with heart failure, arrhythmia, and sudden cardiac death. Accurate diagnosis depends on precise measurement of left ventricular wall thickness across spatial slices and temporal frames of the cardiac cycle. In routine practice, this requires experts to manually delineate endocardial and epicardial boundaries frame by frame, a process that is prohibitively time-consuming and unsuitable for large-scale or real-time deployment. For benchmarking, we used the Sunnybrook Cardiac MRI dataset~\cite{radau2009evaluation}, which consist of 45 cine-MRI images from a mixed of patients and pathologies: healthy, hypertrophy, heart failure with infarction and heart failure without infarction.

In our experiment, TLAgent searched the authoritative clinical definition of hypertrophy, specified as a maximal left ventricular wall thickness $\geq 15$~mm at end-diastole in the absence of secondary causes~\cite{accHCMguideline,aseCMRstandard}. As illustrated in \textbf{Figure~\ref{fig:4}g}, the system assembled a workflow that segmented the left ventricular myocardium (LVMYO), left ventricular blood pool (LVBP), and right ventricular blood pool (RVBP). From these segmentations, endocardial contours were derived from the LVBP boundary, epicardial contours from the outer boundary of LVBP and LVMYO regions, and RV insertion points from RVBP-LVMYO junctions. Slice-wise wall thickness was then computed in 2D by measuring the distance from each endocardial point to its intersection with the epicardial contour along the outward normal direction. These measurements were aggregated across frames to reconstruct dynamic wall-thickness curves. Representative outputs are shown in \textbf{Figure~\ref{fig:4}h}, which display 3D segmentations across multiple time frames together with short-axis 2D overlays on raw cine MRI slices, highlighting the practical utility of TissueLab by providing clinicians with transparent and directly interpretable visual results. The full computational procedure for wall-thickness measurement generated by TLAgent is summarized in Algorithm~\ref{appendix:lvhypertrophy}.

With TLAgent, the automatic design and implementation of this workflow reached an accuracy of 84.2\%, F1 = 0.880, and a strong correlation with ground truth thickness (MCC = 0.701). In contrast, baseline models lack direct 3D data support. To ensure fair comparisons, we manually slice volumetric data into 2D images, specifying the temporal assignment of each slice, and repeating each experiment five times to ensure consistency. The best-performing baseline, GPT-4o-vision, achieved only 57.1\% accuracy and F1 = 0.727, with near-zero correlation to the ground truth, indicating predictions that failed to capture clinically defined patterns. GPT-4o-agent reached F1 = 0.809 but collapsed to predicting nearly all cases as positive, eliminating specificity. LLaVA-Med’s performance was essentially random (accuracy = 57.6\%, F1 = 67.3\%, MCC = 0.089), and MedGemma also showed limited performance (accuracy = 42.9\%, F1 = 0.455). These comparisons emphasize that only TLAgent was exhibited high accuracy with non-trivial recall, sustaining robust performance in a challenging sub-task where vision-language baselines degraded to chance-level behavior.

~\begin{algorithm}
\caption{Pseudocode for LV hypertrophy detection generated by TLAgent}
\label{appendix:lvhypertrophy}
\begin{algorithmic}[1]
\STATE \textbf{Input:} LV blood pool mask $M_{\text{LVBP}}$, LV myocardium mask $M_{\text{LVMYO}}$, RV blood pool mask $M_{\text{RVBP}}$, Pixel spacing $(\Delta x,\Delta y)$, Threshold $\tau = 15$ mm
\STATE \textbf{Initialize:} $WT_{\max}^{\mathrm{ED}} = 0$
\STATE Identify ED frame $t_{\mathrm{ED}}$ \COMMENTT{Frame with maximal LVBP area}
\FOR{each slice $z$ in $t_{\mathrm{ED}}$}
    \STATE $C_{\text{endo}} \gets \partial(M_{\text{LVBP}}[t_{\mathrm{ED}},z])$ \COMMENTT{Endocardial contour}
    \STATE $C_{\text{epi}} \gets \partial(M_{\text{LVBP}}[t_{\mathrm{ED}},z] \cup M_{\text{LVMYO}}[t_{\mathrm{ED}},z])$ \COMMENTT{Epicardial contour}
    \FOR{each $p \in C_{\text{endo}}$}
        \STATE $\mathbf{n}(p) \gets$ outward normal at $p$
        \STATE $q \gets$ first intersection of ray $(p,\mathbf{n}(p))$ with $C_{\text{epi}}$
        \STATE $d \gets \|p-q\| \times \text{PixelScale}(\Delta x,\Delta y)$ \COMMENTT{Local thickness (mm)}
        \IF{$d > WT_{\max}^{\mathrm{ED}}$}
            \STATE $WT_{\max}^{\mathrm{ED}} \gets d$
        \ENDIF
    \ENDFOR
\ENDFOR
\IF{$WT_{\max}^{\mathrm{ED}} \geq \tau$}
    \STATE Hypertrophy $\gets$ Yes \COMMENTT{Guideline-aligned diagnosis}
\ELSE
    \STATE Hypertrophy $\gets$ No
\ENDIF
\STATE \textbf{Output:} Hypertrophy label, $WT_{\max}^{\mathrm{ED}}$
\end{algorithmic}
\end{algorithm}

\subsection{TissueLab interactive ecosystem enables customized tool development}
\label{sec:Co-evolving with clinician-in-the-loop improves the quantification of important tissue/cell for cancer research}

Among above experiments, it is worth noting that TLAgent performance is bottlenecked by underlying tools, such as segmentation model, rather than in the agent orchestration strategy itself.
Although the system can seamlessly integrate improved segmentation tools in the future to further boost accuracy while preserving interpretability and guideline alignment, in reality there are limited tools available for medical imaging analysis, and many research and clinical questions fall into edge cases for which no existing tool can provide solutions.

To address this limitation, TLAgent is deeply integrated into our TissueLab software and ecosystem (\href{https://www.tissuelab.org?from=arxiv}{\textcolor{tissuelabblue}{tissuelab.org}}). The TissueLab platform enables users to create customized tools by (i) developing Python pipelines that can be uploaded to the ecosystem; and (ii) fine-tuning existing foundation models for diverse downstream tasks. Once built, these models can be shared through the TissueLab ecosystem, fostering collaboration and reuse. This capability allows the TissueLab agentic AI system to substantially accelerate medical imaging research and scientific discovery (\textbf{Figure~\ref{fig:5}a}). We evaluated this capability of cell identification on two different cancer pathology slides (Colon and Prostate) from the Visium HD dataset~\cite{10xvisiumhd, oliveira_high-definition_2025}, which includes high-resolution spatial transcriptomics to serve as ground-truth labels.

In this experiment, the objective was to determine how many cells are tumor. Despite our best efforts to ensure fair comparisons, baseline models consistently failed to produce valid outputs for this task, as they may not have had access to the same high-resolution data as TLAgent. However, guided by clinician's active learning annotation, TissueLab was able to evolve from unseen cases toward refined clinical alignment. By allowing pathologist to finetune our in-house \HandE~cell classification foundation model ``NuClass'', the system rapidly improved tumor cell identification performance from near-zero accuracy to 82.1\% within five minutes of incremental training, and ultimately reached 94.9\% accuracy after 30 minutes (\textbf{Figure~\ref{fig:5}b--c}). At the same time, \textbf{Figure~\ref{fig:5}d--e} show that predictions not only matched overall counts but also reproduced the spatial distribution of neoplastic versus non-neoplastic cells, closely aligning with ground truth labels. The intersection-over-union (IoU) between predicted and ground-truth tumor regions improved dramatically from 10.1\% at the initial iteration to 88.9\% in 30 minutes (\textbf{Figure~\ref{fig:5}f}), highlighting TissueLab's capability to enable users to efficiently build a cell-counting model from scratch.

\begin{figure}[tbp]
  \centering
  \includegraphics[width=\textwidth]{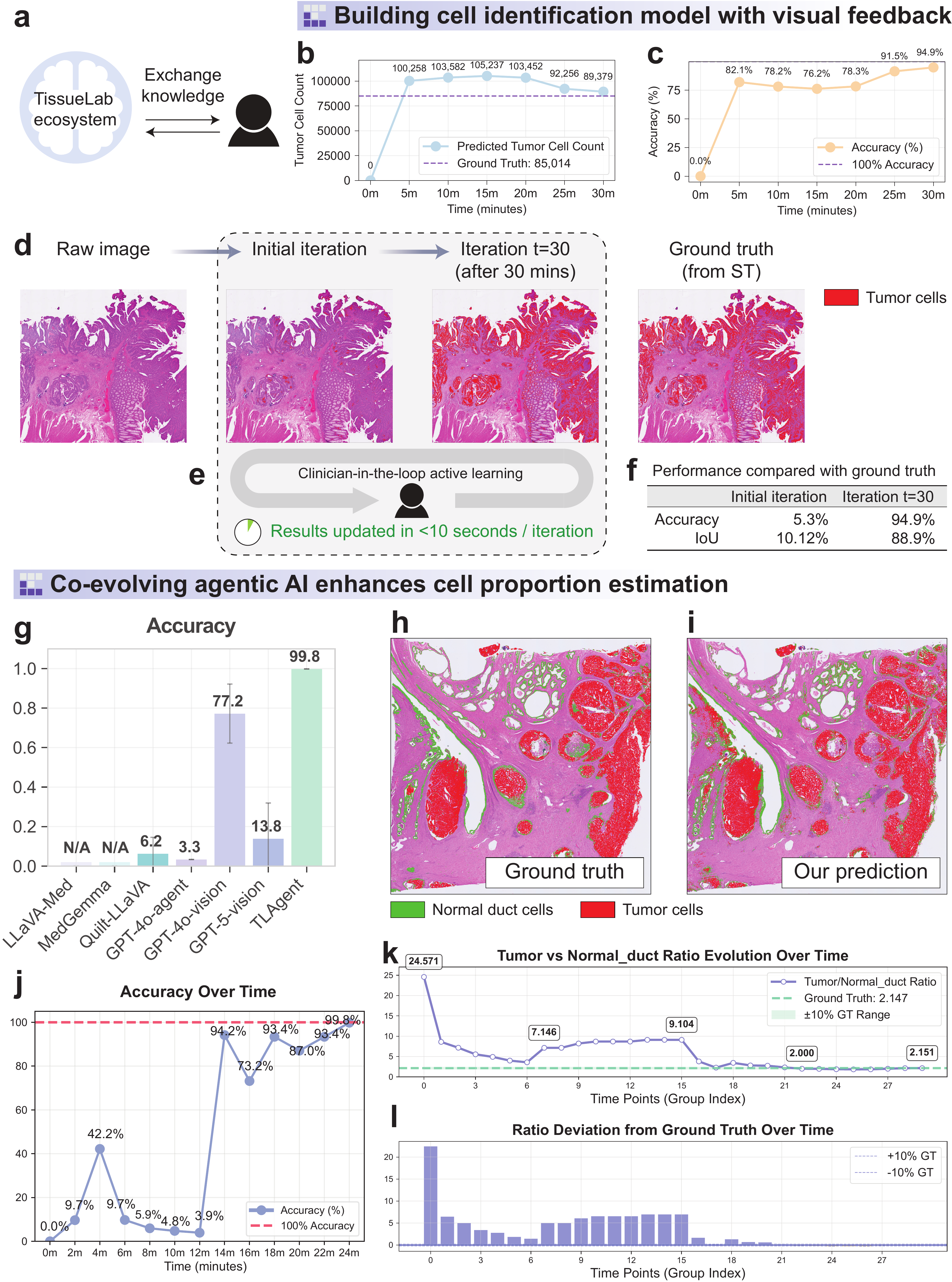}
\end{figure}

\afterpage{%
  \captionsetup{aboveskip=-10pt, belowskip=20pt}
  \captionof{figure}{\small \textbf{Clinician-in-the-loop co-evolution improves tissue and cell quantification in cancer research.}
    \textbf{a}, Schematic illustration of the clinician-in-the-loop mechanism within the TissueLab ecosystem. 
    \textbf{b}, Tumor cell counts over time compared with ground truth (GT). 
    \textbf{c}, Tumor cell detection accuracy as a function of feedback iterations. 
    \textbf{d}, Representative raw image. 
    \textbf{e}, Image series showing progressive improvement in tumor cell detection after iterative feedback. 
    \textbf{f}, Performance of detection compared with GT. 
    \textbf{g}, Comparison of model accuracy in quantifying tumor-to-normal duct cell ratios across methods. 
    \textbf{h}, Cell overlay from ground truth. 
    \textbf{i}, Cell Overlay of TissueLab predictions.
    \textbf{j}, Accuracy of tumor-to-normal duct cell ratio estimation over time. 
    \textbf{k}, Evolution of predicted tumor-to-normal duct cell ratios across feedback iterations. 
    \textbf{l}, Deviation from GT over time.}
  \label{fig:5}%
}

We next evaluated the ratio of tumor cells to normal duct cells in prostate tissue, a measure with clinical relevance for quantifying malignant transformation, informing prognosis, and supporting cancer staging. By repeating this experiment five times, GPT-4-vision reached 77.2\% accuracy but with excessively wide prediction distributions, reflecting random guesses rather than consistent reasoning; and all other baselines fell below 12\% accuracy or failed outright (\textbf{Figure~\ref{fig:5}g}). In contrast, the TissueLab platform enables clinicians to observe workflow results in real time, track how models learn from their annotations, and co-evolve with the system. Through this process, lightweight classifiers are trained on top of our in-house ``NuClass'' model and shared within the ecosystem. By applying the clinician-guided classifiers, TLAgent achieved 99.8\% accuracy, and as shown in \textbf{Figure~\ref{fig:5}h--i}, the predicted tumor and duct distributions closely matched the ground truth spatial patterns.

Additionally, improvements in accuracy, cell ratio, and ratio deviation over time are presented in \textbf{Figure~\ref{fig:5}j--l}, respectively. Because expert annotations involved labeling multiple cell types, the ratio initially showed little improvement. However, once the relevant cells were incorporated, TissueLab rapidly adapted: accuracy rose from 3.9\% to above 90\% within two minutes of feedback, and continued to improve toward near-perfect performance, ultimately reaching 99.8\%. At the same time,  the predicted ratios evolved smoothly toward the ground truth with minimal deviation across iterations.

These findings confirm that co-evolving with human-in-the-loop component through TissueLab platform enables user to iteratively refine and customize existing models to achieve expert-level performances. With human expert supervision, the system uses active learning to adapt from novel cases to near-perfect concordance within minutes, offering immediate research and clinical utility. Such capability requires an interactive platform that enables timely, bidirectional feedback between human experts and the AI system, a capability not attainable with current VLM-based models or other agent systems.

\subsection{TissueLab enables multi-modal integration of spatial omics and histology to improve accuracy in pathology analysis}
Beyond achieving superior performances over baseline models on both pathology and radiology tasks, the TissueLab platform and TLAgent also demonstrate the ability to perform multi-modal integrative analysis. Here we demostrate a use case where TLAgent can recognize the presense of both spatial omics and histopathology modalities thereby generating more precise and clinically meaningful tissue analysis workflow~\cite{rao_multimodal_2025}. Specifically, we evaluate this capability in the context of glomerulus quantification on a kidney whole slide image from the Visium HD dataset, which provides high-resolution spatial transcriptomics data. 
As a common quantification task, accurate glomerulus counting is critical for nephrology research and clinical care, particularly for advanced measures such as the fraction of globally sclerotic glomeruli. However, these assessments have traditionally relied on labor-intensive manual evaluation by expert pathologists.

\begin{figure}[hbtp]
    \centering
    \includegraphics[width = \textwidth]{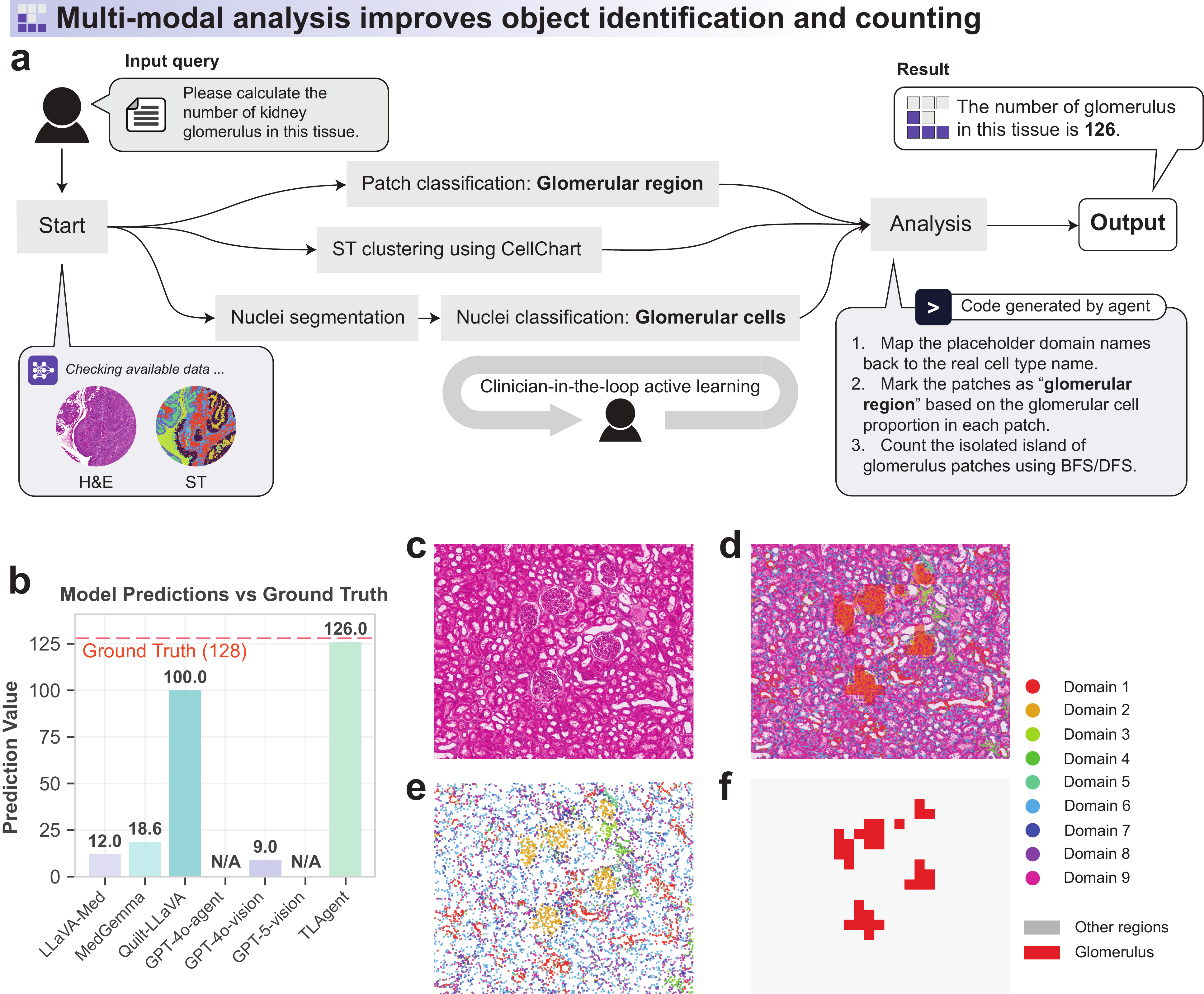}
    \caption{\small \textbf{Multi-modal analysis integrating spatial omics and histopathology for glomerulus quantification.} 
    \textbf{a}, Schematic diagram of the TissueLab multi-modal analysis workflow. 
    \textbf{b}, Comparison of model-predicted versus ground-truth (GT) glomerulus counts. 
    \textbf{c}, Representative kidney \HandE~image. 
    \textbf{d}, Patch-level and cell-level result overlays. 
    \textbf{e}, Spatial domain clustering result obtained by integrating spatial omics data with histology. 
    \textbf{f}, Glomerulus region prediction result.}
    \label{fig:6}
\end{figure}

TLAgent accomplished this task through a structured multi-modal workflow (\textbf{Figure~\ref{fig:6}a}). First, the system will check available data in the data container. If both the \HandE~and spatial transcriptomics data are available, it will performed cell segmentation on \HandE~images and extracted the centroids of individual cells. These spatial coordinates were then linked to gene expression profiles, and unsupervised clustering was performed to assign each cell to a transcriptomic domain.  Because clustering yields only numerical domain labels without inherent biological meaning, TLAgent established multi-modal correspondence by mapping these domains back to annotated \HandE~regions. Through this alignment, the cluster corresponding to glomerular cells was identified. Once identified, TLAgent computed the local proportions of glomerular versus non-glomerular cells, aggregated glomerulus patches, and applied graph-based traversal (depth-first and breadth-first search, DFS/BFS) to obtain discrete glomerulus counts. This end-to-end workflow autonomously generated by TLAgent, highlights both the reasoning capabilities of LLMs and the local data access features embedded in our design.

With an appropriate domain configuration, TLAgent achieved 98.4\% accuracy, nearly indistinguishable from expert annotations (\textbf{Figure~\ref{fig:6}b}). In contrast (\textbf{Figure~\ref{fig:6}b}), although we have made every effort to ensure fair comparisons with other baseline models and systems, baseline vision-language models were still unable to cope with the scale and complexity of spatial omics data, and consistently failed to complete the task, producing outputs that were either random or absent. Through TissueLab platform's visualization interface (\textbf{Figure~\ref{fig:6}c--f}), clinicians can directly inspect overlays of predictions with histological images, confirming that the identified glomerulus regions and counts are highly consistent with expert assessments.

These results demonstrate that TissueLab can still deliver expert-level performance in a setting where multi-modal data are available. By integrating molecular and morphological information, the system not only provides richer insights into kidney pathology but also establishes a scalable framework for multi-modal medical AI analysis.

\section{Discussion}
\label{sec:discussion}
We bring the concept of ``\textbf{laboratory intelligence}'', devoting extensive engineering and design efforts in building TissueLab ecosystem and TLAgent and making them openly accessible to the community. TissueLab is fully open source, with installers for Windows, macOS, and Linux, and a freely accessible web portal (\href{https://www.tissuelab.org?from=arxiv}{\textcolor{tissuelabblue}{tissuelab.org}}). As one of the first agentic AI systems aimed at translating advances in medical imaging AI research into tools readily applicable to both day-to-day research and clinical practice, the TissueLab ecosystem and TLAgent offer a modular and extensible framework that enables clinicians, physician-scientists, lab technicians, and AI researchers to directly integrate state-of-the-art AI methods into their workflows. In doing so, TissueLab bridges the gap between algorithmic innovation and clinical/translational adoption, functioning as a programmable laboratory for medical imaging with modular, reconfigurable infrastructures that lower entry barriers and accelerate scientific discovery.

The strong performance of TissueLab is largely attributable to its unique ability to co-evolve with clinicians. The system allows user to insert a new or finetune an existing model, thereby rapidly adapt to new diseases with only minutes of annotation, an important feature that is critical for many research and clinical settings where AI can benefit largely on those difficult diesases that are rare to collect training data. Moreover, by establishing a shared ecosystem, TissueLab enables collaborative knowledge transfer across diverse tasks and base models. In this way, the system evolves from unseen cases to align with clinical demands, ensuring that each new interaction enhances its overall clinical intelligence.

Compared with prior foundation models and some other vision-language-based agentic AI systems, which often lack interpretability therefore prevents adoption in clinical setting, TissueLab is explicitly well-suited for translational applications. Most existing agentic AI systems operate as black boxes~\cite{ferber_development_2025}, offering limited insights into the biological mechanisms behind their predictions, which makes them difficult to trust and deploy in high-stakes medical contexts~\cite{amann2020explainability}. In contrast, TissueLab and TLAgent makes all intermediate data fully visualizable, allowing clinicians to inspect and interpret the reasoning process. Moreover, through its evolving memory layer, all intermediate outputs are not only transparent e but also editable, enabling users to directly trace, adjust, and fine-tune the reasoning process from iterative interaction (the ``co-evolving'' feature). In addition, TissueLab leverages the Model Context Protocol (MCP) to retrieve diagnosis in the latest and authoritative clinical guidelines, and because such guidelines evolve over time, this dynamic knowledge-retrieval ensures the system remains sustainable and aligned with the latest standards of care. This design ensures that the system remains transparent and continuously improves with minimal effort, while keeping clinicians in full control of the decision pathway. Furthermore, TissueLab supports the seamless integration of new models into its modular framework, thereby maintaining state-of-the-art performance while ensuring transparency and clinical alignment.

Through the online TissueLab ecosystem, we anticipate broader use of TLAgent, with user-provided knowledge, annotation, and model contributions continually enhancing its intelligence.

Despite these promising results, our study has several limitations. First, while there has been recent progress in self-evolving AI~\cite{kuba_language_2025, yuan_self-rewarding_2025, li_self-rewarding_2025, sun_seagent_2025}, TissueLab currently co-evolves with clinicians in the loop to ensure its human-aligned safety. This design minimizes risks but also increases the level of human supervision required. In the future, we hope to move toward a community-driven, self-evolving autonomous AI co-scientist that can continuously improve under established safety constraints while reducing dependence on manual supervision.
Second, although we have made every effort to ensure fair comparisons with other baseline models and systems, it is important to note that some baselines may not have access to the same high-resolution data (such as gigapixel pathology images) available to TLAgent. For instance, the pathology input to GPT-5-vision is restricted to the largest thumbnail the model can accept, whereas TLAgent operates directly on raw whole-slide images. We anticipate that future work could develop a GPT-5-vision-based pipeline for fairer comparison, though building such a system would itself require substantial effort.
Lastly, integrating TissueLab into routine clinical workflows will require not only technical validation but also compliance with regulatory standards and interoperability with existing hospital systems. Addressing these limitations will be critical for translating TissueLab into a widely adopted and safe clinical intelligence system.

These limitations also motivate future work and underscore the potential of agentic AI systems to progress from ``laboratory intelligence'' towards ``clinical intelligence''. A natural next step is to further advance TissueLab into a community-driven self-evolving ecosystem~\cite{ritore_open_2024, wolf_huggingfaces_2020}, where contributions from the broader TissueLab community collectively guide its evolution. By pooling diverse expertise and data within a controlled and transparent framework, TissueLab could continuously refine its capabilities by self-evolving, adapt more rapidly to emerging clinical challenges, and reduce reliance on manual oversight.

\section{Methods}
\label{sec:methods}
\subsection*{Adaptive and extensible agentic system for seamless model integration}

The field of medical agentic AI is advancing rapidly, with successive generations of agents frequently exceeding the capabilities of their predecessors. Yet much of this apparent progress has come from continually replacing or updating the underlying tool models---a costly and ultimately unsustainable strategy. True evolution requires more than incremental tool updates; it demands the development of agentic AI systems that are inherently extensible and capable of self-adaptation.  

To address these needs, TissueLab was designed as an adaptive and extensible framework. TissueLab agents can seamlessly integrate \emph{any} model as it becomes available, without requiring modifications to the orchestration logic or reliance on a fixed toolbox. Mathematically, let $\mathcal{M}_t^{\mathrm{any}}$ denote the set of all models integrated at time $t$. The effective capability of the system can be expressed as
\[
\mathrm{Capability}(t) \;\propto\; \sum_{m \in \mathcal{M}_t^{\mathrm{any}}} \mathrm{Perf}(m),
\]
highlighting that system capacity expands monotonically with the integration of arbitrary models from the broader medical AI ecosystem.  

Adaptivity in TissueLab is realized through a modular plugin architecture, inspired by design patterns in software engineering---particularly the \emph{factory method}---that emphasize modularity and separation of concerns. Computational models are abstracted as task nodes within a directed acyclic workflow graph, making it straightforward to expand the system. Each node exposes a standardized interface with three essential functions: initialization to declare requirements and parameters, input handling to parse data into the expected format, and execution to perform model computation. The internal implementation of each model remains a black box to the agent, enabling uniform orchestration regardless of modality or complexity. This abstraction ensures that any state-of-the-art model can be mounted as a new node, allowing TissueLab to continuously incorporate advances in medical AI while preserving scalability and robustness.  

\subsection*{Topological sorting enables parallel and distributed inference}
TissueLab organizes computational models as modular task nodes within a directed acyclic workflow graph. By applying topological sorting~\cite{kahn1962}, the system ensures that tasks with dependencies (e.g., segmentation before classification) are executed in sequence, while independent branches can be run in parallel. This design improves scalability and reduces inference time, and also allows distributed execution across multiple GPU servers.

Importantly, dependencies are defined only in a pairwise manner (e.g., segmentation precedes classification), which avoids the complexity of reasoning over a full graph and reduces the likelihood of hallucinations by the LLM. This guarantees that clinically meaningful workflows remain both efficient and trustworthy, even in large-scale imaging studies. When executing planned workflows, the agent does not need to reason about the complete set of dependency relations among all models, which could otherwise increase complexity and risk hallucinations. Instead, only pairwise dependencies need to be specified; for example, segmentation should always precede classification. Based on this set of pairwise relations, the system performs a topological sorting to construct the workflow. The algorithm counts the in-degree of each task node, begins execution from nodes with zero in-degree, and after each node finishes, removes it from the graph while updating the in-degree of its successors. This procedure enables task nodes to be executed in parallel while preserving the logical order of dependencies. This design naturally accommodates both serial and parallel execution: tasks with strict dependencies (e.g., segmentation preceding classification) are executed sequentially, while independent branches can run concurrently to improve efficiency.

Formally, let the workflow be represented as a directed acyclic graph 
$G=(V,E)$ with in-degree function $d(v)$. At each iteration, all nodes 
$v \in V$ with $d(v)=0$ are executed, then removed from $G$, and 
$d(u)$ is decremented for all $(v,u)\in E$:

\[
\forall v \in V, \quad 
d(v) = 0 \;\;\Rightarrow\;\; 
\text{execute}(v),\; V \leftarrow V \setminus \{v\},\; 
d(u) \leftarrow d(u)-1 \;\;\forall (v,u)\in E.
\] 

\noindent The procedure can equivalently be expressed as pseudocode:

\begin{algorithm}
\caption{Parallel Topological Sort}
\label{alg:parallel-toposort}
\begin{algorithmic}[1]
\STATE \textbf{Input:} Directed acyclic graph $G = (V, E)$
\WHILE{$V$ is not empty}
    \STATE $S \gets \{ v \in V \mid indegree(v) = 0 \}$
    \STATE \textbf{execute all $v \in S$ in parallel}
    \FOR{each $v \in S$}
        \STATE remove $v$ from $V$
        \FOR{each $(v, u) \in E$}
            \STATE $indegree(u) \gets indegree(u) - 1$
        \ENDFOR
    \ENDFOR
\ENDWHILE
\end{algorithmic}
\end{algorithm}

This procedure ensures that workflow execution remains deterministic and scalable: deterministic because the topological order is uniquely determined by the dependency graph, and scalable because independent branches of the workflow can be executed in parallel without violating dependencies. By reducing the need to reason over the entire dependency structure, the design also improves robustness and lowers the risk of hallucinations during orchestration.

\subsection*{Community-driven co-evolution}
The co-evolving mechanism in TissueLab consists of two complementary components. First, human-in-the-loop feedback (e.g., clinician corrections on image patch-level classification) is captured as additional labeled data and incorporated into an active learning loop. These annotations are used to update lightweight modules (e.g., shallow classifiers such as XGBoost~\cite{chen_xgboost_2016}), thereby improving reliability and clinical alignment. Because such modules can be retrained within minutes using a small number of user-provided annotations, the system can rapidly adapt to specific clinical scenarios without relying on large-scale pretraining or extensive datasets. Formally, if $\theta$ denotes the parameters of a lightweight model, then feedback data $\{(x_i, y_i)\}$ is used to solve
\[
\theta^\star = \arg\min_{\theta} \sum_i \ell\!\left(y_i,\, f_{\theta}(x_i)\right),
\]
where $\ell$ is the training loss and $f_\theta$ is the lightweight model. This update mechanism enables rapid personalization and clinician-aligned adaptation. Importantly, within the TissueLab ecosystem, these adapted modules can be shared across the community or further refined collaboratively, enabling collective improvement of classifiers and fostering continuous system-wide progress. In this way, contributions from individual users directly benefit the broader community.  

Second, the agent maintains an extensible candidate pool of available models, which can incorporate any new models from the broader medical AI ecosystem on demand. Model selection or switching is driven by user evaluation of the outputs (e.g., whether results are deemed acceptable or inadequate), rather than performed autonomously. Formally, a large language model first retrieves a task-relevant candidate set $\mathcal{M}_t^{\mathrm{cand}} \subseteq \mathcal{M}_t^{\mathrm{any}}$. A feedback-guided policy $\pi_{\phi}$, trained on accumulated feedback $\mathcal{F}_t$, then assigns rank scores to models in this candidate set:
\[
s_m = \pi_{\phi}(m \mid \mathcal{M}_t^{\mathrm{cand}}, \mathcal{F}_t),
\qquad 
\mathcal{M}_t^\star = \mathrm{Rank}\!\big(\mathcal{M}_t^{\mathrm{cand}}, \{s_m\}\big).
\]
This design ensures that model selection adapts dynamically with user feedback while leaving outputs unaltered. Furthermore, TissueLab provides a community-driven mechanism whereby AI researchers can seamlessly integrate their latest models into the agentic system, making cutting-edge research immediately available to clinicians in real workflows. This not only accelerates the translation of AI research into practice but also ensures that advances from the evolving AI ecosystem are leveraged safely, under explicit human oversight.

\subsection*{LLM orchestration}
TissueLab is built upon a large language model (LLM) backbone. In our experiments, we employ chatgpt-4o-latest (OpenAI, 2025) as a representative implementation, though any future LLM can serve this role. The LLM is used only as an orchestration layer: it plans workflows, generates code, and invokes function calls, but it does not directly process raw medical images. All data analysis tasks (e.g., segmentation, classification, code analysis) are executed locally by domain-specific models and pipeline. This design avoids token overload and attention dilution when handling gigapixel whole-slide images or volumetric radiology data, while enabling the system to remain adaptive and extensible. Importantly, the modular design allows the base LLM to be replaced by future models without modifying the orchestration framework, making the system adaptive and future-proof.

\subsection*{Execution layer: interaction between agent and tool models}
 The execution layer consists of domain-specific models and pipelines, which are invoked as task nodes by the agent. all intermediate data states are persisted in a shared local storage (HDF5 database~\cite{folk_hdf5_2011}), rather than being passed directly between tools. Each tool model is treated as a black box task node that exposes a standardized read() interface to retrieve the required inputs from the shared database, without needing to be aware of upstream or downstream dependencies. The orchestration layer (LLM) ensures that, at the time of execution, all necessary inputs for a given node have already been generated and stored. This design decouples tools from one another, simplifies integration of new models, and guarantees reproducibility by making all intermediate outputs explicitly available in the persistent storage.

\subsection*{Editable memory layer}
To address the challenges of token overload and attention dilution~\cite{liu_lost_2024} in long multi-step workflows, TissueLab implements an editable memory layer. Unlike conventional LLM-based systems that rely solely on limited token context, TissueLab employs persistent memory implemented through local structured storage (HDF5 databases). All intermediate results, annotations, and execution logs are written into this memory layer, which can be accessed by downstream task nodes via standardized interfaces. Because the memory resides outside of the token context, its capacity is effectively limited only by hardware storage rather than by model constraints, enabling theoretically unbounded memory.

Importantly, the memory layer is fully visible to users. Clinicians or researchers can visualize the result of each step within the TissueLab platform for review, and they can refine performance by augmenting stored results and annotations. Such updates are directly incorporated into subsequent executions, thereby improving the agent’s performance over time. By storing explicit intermediate outputs that are both persistent and editable, the memory layer ensures that workflows remain transparent, auditable, and adaptable, with every decision traceable to concrete stored data rather than transient model states.

\subsection*{Semantic function-calling for data access}
The Data Access Layer employs function calling to read the structure and content of locally stored HDF5 files. All datasets follow a semantic naming convention (e.g., \texttt{mask2D}, \texttt{volume3D}, \texttt{3Dmask\_\{timestamp\}}, \texttt{4Dseries\_\{phase\}}), and carry metadata such as \texttt{voxel\_spacing}, \texttt{origin}, \texttt{orientation}, and \texttt{timestamp}. By inspecting both the dataset name and its structure, the LLM can infer the role of the data-for example, a dataset named \texttt{3Dmask} with three dimensions is recognized as a volumetric segmentation mask.  

When a task requires specific inputs, the LLM issues function calls that query the HDF5 structure, retrieves the identified datasets together with their metadata, and generates local code to run the analysis pipeline. Computation is thus grounded in concrete structured data of arbitrary dimensionality, rather than within the token context of the model, avoiding attention dilution and token overload that commonly arise with gigapixel slides or volumetric scans.  

Because access is explicit and aligned with the HDF5 schema, outputs are reproducible, auditable, and privacy-preserving, as sensitive medical data remain local. Extending the system to new modalities or dimensions only requires adding semantically named datasets, without modifying the orchestration logic, thereby further strengthening adaptivity. Together with the Editable Memory Layer, this design provides both persistent storage and semantically structured access, forming a reliable foundation for building co-evolving agentic AI workflows.

\subsection*{MCP for searching external criteria}
In the medical domain, these challenges are further compounded by the necessity of clinical oversight: reliance on LLM-generated reasoning alone is insufficient, as clinical expertise remains indispensable for validating, correcting, and guiding AI outputs. Only through such integration can agentic systems mitigate hallucinations and ensure that their recommendations remain safe, reliable, and aligned with established standards of care. many tasks require reference to established guidelines or criteria, such as cancer staging systems, diagnostic thresholds, or treatment protocols. To ensure that agentic workflows are grounded in authoritative sources, TissueLab employs the Model Context Protocol (MCP) exclusively for external knowledge retrieval. MCP is not used for communication between local tool models, nor does the system rely on the LLM to guess criteria or on a fixed internal knowledge base. Instead, MCP treats the entire web as a knowledge repository, allowing the agent to continuously query up-to-date standards and guidelines as they evolve.  

When the agent encounters a task that requires domain-specific thresholds or criteria (e.g., tumor staging according to AJCC, or risk stratification following TCGA-defined biomarkers), it issues an MCP query and incorporates the retrieved information into the workflow. This design guarantees that outputs are supported by traceable references, thereby reducing the risk of hallucination and enhancing clinical trust.  

By restricting MCP to the retrieval of external evidence, TissueLab ensures that local computation remains efficient and privacy-preserving, while clinical outputs remain verifiable and aligned with evolving medical standards.

\subsection*{Benchmarking TissueLab agentic AI system on diverse clinical imaging cohorts}
We evaluated TissueLab across diverse datasets spanning pathology, radiology, and spatial omics. Pathology tasks included whole-slide images of regional lymph node metastasis in colon adenocarcinoma (LNCO2 dataset, Region Gävleborg Clinical Pathology and Cytology department) and prostate and colon cancer slides from VisiumHD. Radiology experiments were performed on multiple public cohorts, including fatty liver assessment (Kaggle: UNIFESP Chest CT Fatty Liver Competition), thoracic disease detection from chest X-rays (Kaggle: NIH Chest X-rays), cardiac hypertrophy analysis from the Sunnybrook Cardiac Data (SCD), and intracranial hemorrhage detection from the PhysioNet Computed Tomography Images for Intracranial Hemorrhage dataset. For spatial omics, we used VisiumHD kidney tissue samples, enabling cross-modal integration with corresponding \HandE~sections.

In the experiments related to LNCO2, we carefully constructed non-overlapping cohorts. Depth of invasion analysis was performed on 107 primary tumor site slides, while positive lymph node counting used 321 lymph node slides, with the two cohorts defined as mutually exclusive (428 slides excluded in total). For independent evaluation, we reserved a separate hold-out set of 50 slides from 6 patients, which were provided to pathologists for detailed annotation of tumor epithelium, stroma, and other compartments. To minimize potential confounding, these hold-out cases were chosen from samples not involved in any of the other experiments.

For fatty liver diagnosis, the dataset comprises 152 patients with chest CT scans (CS) and 112 patients with abdomen or combined abdomen-chest CT scans (AS), for a total of 264 patients. The data were split at the patient level into training and testing cohorts. The training set included 226 patients (114 CS and all 112 AS), while the testing set comprised the remaining 38 CS patients.

For VisiumHD, we directly applied TissueLab’s co-evolving framework, enabling clinicians to rapidly obtain new task-specific models from scratch through interactive feedback, without requiring large-scale pretraining.

To contextualize performance, we benchmarked TissueLab against state-of-the-art vision-language and agentic AI baselines. To ensure a fair comparison, all experiments were repeated five times, and mean values with standard deviations (mean~$\pm$~s.d.) are reported; error bars in figures correspond to these standard deviations. For pathology, comparisons included LLaVA-Med~\cite{li_llava-med_nodate}, MedGemma~\cite{sellergren_medgemma_2025}, Quilt-LLaVA~\cite{seyfioglu_quilt-llava_2025}, GPT-4o-agent, GPT-4o-vision, and GPT-5-vision. For radiology, benchmarks included LLaVA-Med, MedGemma, M3D~\cite{bai_m3d_2024}, GPT-4o-agent, GPT-4o-vision, and GPT-5-vision. In the case of spatial omics integration, existing VLM- and LLM-based agents lack the capacity to align gene expression data with \HandE~images and cannot handle high-dimensional expression matrices that exceed their token limits. As such, only comparisons with ground truth were possible in this domain.

In conclusion, these datasets cover a broad spectrum of tissue types, imaging modalities, and clinical tasks, providing a comprehensive benchmark for assessing the adaptability of our agentic AI system.

\subsection*{Tissue segmentation and classification}
In pathology, there is currently no single universal model capable of performing accurate tissue segmentation and classification across all organ systems. For example, some models perform well for colorectal cancer slides but fail to segment and classify structures such as glomeruli in kidney biopsies. To address this limitation, TissueLab adopts a patch-level segmentation and classification strategy. Before patch-level analysis, TissueLab crop the Whole-slide images (WSIs) into multiple patches. Then perform background removal to ensure that only relevant tissue regions are retained. Specifically, low-resolution slide thumbnails are first processed with adaptive thresholding to distinguish tissue from background under varying staining and illumination conditions. The resulting masks are then refined using morphological closing and hole-filling operations to repair broken regions and fill internal gaps, followed by connected-component analysis to filter out small noisy fragments. Finally, boundary constraints are applied to remove common edge shadows and artifacts near the slide borders. This step-wise procedure progressively cleans artifacts, enhances mask continuity, and supports saving of intermediate results for debugging and visualization. Together, these operations yield accurate and stable tissue masks, ensuring that subsequent patch extraction and downstream analyses focus on true tissue regions rather than background noise. After preprocessing, TissueLab uses foundation models such as MUSK~\cite{xiang_visionlanguage_2025} or PLIP~\cite{huang_visuallanguage_2023} to extract embeddings from tissue patches. Because these embeddings are trained with contrastive learning to capture transferable semantic representations, they can be applied in a zero-shot learning paradigm for initial segmentation and classification.

When the zero-shot predictions are insufficient, users can provide patch-level annotations that serve as feedback to the agentic system. Such annotations enable rapid retraining of lightweight models (e.g., linear heads) to specialize for the current task, often within minutes. Additionally, other tissue segmentation or tissue classification models developed by other researchers can be seamlessly integrated through the TissueLab ecosystem, further enhancing performance in domain-specific tasks. This multi-source and feedback-driven design allows the system to progressively improve agentic AI's accuracy in a co-evolving manner, while remaining adaptable to diverse tissue types and pathological contexts.

In contrast to pathology, the radiology domain already offers a broad range of high-performance pretrained models for segmentation and analysis. TissueLab leverages this ecosystem by integrating domain-specific models into its model factory as task nodes. Any pretrained radiology model can be registered in the factory. When a new query arrives, the LLM dynamically selects a context-relevant candidate pool for orchestration:
\[
\mathcal{M}_t^{\mathrm{cand}} = \mathrm{LLM}\!\left(\text{query}, \,\mathcal{M}_t^{\mathrm{factory}}\right).
\]
This design ensures that only appropriate domain-specific models are considered for execution, reducing unnecessary computation while maintaining flexibility.  

As illustrative examples, TotalSegmentator~\cite{wasserthal_totalsegmentator_2023} and BiomedParse~\cite{zhao_foundation_2025} are incorporated as task nodes. TotalSegmentator provides fine-grained, high-accuracy segmentation of multiple anatomical structures across CT volumes, but is limited to a predefined set of organs and regions exposed through its interface. Meanwhile, BiomedParse enables query-driven segmentation of arbitrary user-specified structures, offering flexibility when the requested target is not part of the TotalSegmentator set.

This ensures that TissueLab leverages the strengths of the broader radiology AI ecosystem without duplicating effort, functioning as a flexible orchestration framework that can dynamically select domain-specific models to deliver more precise predictions in practical clinical workflows.

\subsection*{Cell segmentation and classification}
To enable cell-level analysis, TissueLab integrates both external and in-house models within its agentic AI framework. For nucleus segmentation, TissueLab incorporates StarDist~\cite{schmidt2018, weigert2020, weigert2022} as a tasknode, an external foundation model that provides polygonal instance segmentation with high accuracy and robustness across whole-slide images (WSIs). To ensure reliable detection of nuclei at patch boundaries, WSIs are processed using overlapping tiles. Overlapping regions are then deduplicated to prevent double counting, thereby maintaining accurate cell counts across the entire slide. This design ensures that segmentation remains scalable to gigapixel pathology images while preserving precision at the single-cell level.

Building on this segmentation backbone, TissueLab also integrates NuClass, an in-house vision--language foundation model designed for nuclei typing under a zero-shot paradigm. NuClass combines three key innovations: it leverages hierarchical biomedical ontologies and natural-language descriptions to enable agents to reason over semantic relationships between cell classes and adapt to unseen types; it adopts a nucleus-centered patch representation that naturally guides transformer attention to biologically meaningful regions, improving the precision of nucleus-level analysis; and it is pretrained on over 8.3 million nuclei patches from 11 diverse datasets, providing robust morphological priors across tissues that TissueLab agents can readily exploit in varied experimental settings.

Through this integration, TissueLab agents can perform scalable and ontology-aware cell classification, enabling downstream biomedical analyses such as cancer research, spatial gene expression studies, and discovery of disease-related cellular patterns.

\subsection*{Multi-modal analysis}
TissueLab agentic AI system enables multi-modal analysis, where molecular signals from spatial omics can be interpreted together with histological features. By combining active learning outputs from cell-level classification with spatial gene expression data, the system provides richer and more comprehensive information for downstream analysis.

To support this workflow, TissueLab integrates CellCharter~\citep{varrone_cellcharter_2024} for the identification, characterization, and comparison of spatial clusters from spatial omics data. While CellCharter focuses exclusively on spatial omics clustering, TissueLab links these clusters to cell-type information derived from NuClass-based classification or user-provided annotations collected through active learning. This correspondence allows users to interpret spatial transcriptomic clusters in terms of underlying histological cell types and biological functions, thereby advancing beyond modality-specific analyses.  

In this way, TissueLab enables a multi-modal mapping between molecular clusters and cellular phenotypes, extending beyond single-modality processing to support multi-scale, multi-modal insights into tissue organization and disease mechanisms.

\subsection*{Evaluation metrics}
We evaluate TissueLab using a broad spectrum of metrics to capture task completion, predictive accuracy, robustness, and clinical relevance.

\textbf{Task-level metrics.}  
We define model successfulness, or Task Completion Rate (TCR), as the proportion of tasks for which the agent produces a concrete and clinically relevant answer. Formally,
\[
\text{TCR} = \frac{N_{\text{success}}}{N_{\text{total}}},
\]
where $N_{\text{success}}$ denotes the number of queries for which the system output directly addresses the posed clinical question, and $N_{\text{total}}$ is the total number of queries. A response is counted as successful only if it goes beyond generic or evasive statements and yields an actionable result that a clinician could use in practice. This metric is particularly critical for evaluating large language model (LLM)-based systems: unlike traditional accuracy measures, model successfulness explicitly penalizes hallucinated or superficial outputs, thereby ensuring that reported answers are both relevant and clinically interpretable. In this way, model successfulness captures the pragmatic utility of the agentic AI system in real-world workflows, rather than its performance on isolated benchmarks.

\textbf{Regression metrics.}  
For continuous prediction tasks, we report the Mean Absolute Error (MAE) and Root Mean Square Error (RMSE). The MAE,
\[
\text{MAE} = \frac{1}{n}\sum_{i=1}^{n}\big|\hat{d}_i - d_i\big|,
\]
quantifies the average absolute deviation between the predicted value $\hat{d}_i$ and the ground truth $d_i$ across $n$ samples. Because MAE is expressed in the same units as the target variable (e.g., millimeters for tumor invasion depth), it provides an intuitive measure of the typical error magnitude. The RMSE,
\[
\text{RMSE} = \sqrt{\frac{1}{n}\sum_{i=1}^{n}(\hat{d}_i - d_i)^2},
\]
applies a quadratic penalty to deviations, making it more sensitive to large errors. This property highlights the occurrence of rare but clinically consequential misestimations, such as substantially underestimating tumor size or lesion volume. Together, MAE captures average-case reliability while RMSE emphasizes the risk of extreme errors, ensuring that both common and catastrophic failures are evaluated in clinical benchmarking.

\textbf{Pearson correlation coefficient.}  
The Pearson correlation coefficient ($r$) measures the strength of the linear association between predictions and ground truth values. It is defined as
\[
r = \frac{\sum_{i=1}^{n}(d_i - \bar{d})(\hat{d}_i - \overline{\hat{d}})}
{\sqrt{\sum_{i=1}^{n}(d_i - \bar{d})^2}\sqrt{\sum_{i=1}^{n}(\hat{d}_i - \overline{\hat{d}})^2}},
\]
where $d_i$ denotes the ground truth, $\hat{d}_i$ the prediction for sample $i$, and $\bar{d}$ and $\overline{\hat{d}}$ are their respective sample means. Values of $r$ range from $-1$ (perfect inverse correlation) to $+1$ (perfect positive correlation), with 0 indicating no linear association. Unlike error-based metrics such as MAE or RMSE, $r$ is invariant to scale and offset, and thus evaluates whether the relative ordering of samples is preserved. This property is especially important in clinical contexts such as risk stratification, where the correct ranking of patients (e.g., higher invasion depth implying higher risk) may be more consequential than the exact numerical estimate.

\textbf{Classification metrics.}  
For discrete prediction tasks, we report a suite of complementary classification metrics. Let $TP$ denote true positives (correctly identified positive cases), $TN$ true negatives (correctly identified negative cases), $FP$ false positives (negative cases incorrectly predicted as positive), and $FN$ false negatives (positive cases incorrectly predicted as negative).  
  
Accuracy is the most widely used classification metric, defined as the proportion of correctly classified samples among all cases:
\[
\text{Accuracy} = \frac{TP + TN}{TP + TN + FP + FN}.
\]
It provides an intuitive single-number summary of overall correctness and is easy to interpret across different tasks. However, accuracy alone can be misleading in settings with severe class imbalance, as a model may achieve deceptively high scores by always predicting the majority class. For instance, in cancer screening with 95\% healthy cases, a trivial model labeling all patients as healthy would still reach 95\% accuracy while completely failing to detect diseased individuals. These limitations motivate the complementary use of precision, recall, and F1-score, which better capture diagnostic reliability under clinical imbalance.
 
To complement accuracy, we report precision, recall, and their harmonic mean (F1-score), which together provide a more nuanced assessment of diagnostic performance under class imbalance. 

Precision quantifies the reliability of positive predictions, indicating the proportion of predicted positives that are truly positive.
\[
\text{Precision} = \frac{TP}{TP + FP}.
\]
A high precision indicates that most predicted positives are true positives, thereby reducing the risk of over-diagnosis and unnecessary clinical interventions.  

Recall (sensitivity) captures the model’s ability to identify all true positive cases, thereby minimizing false negatives. This is crucial in clinical contexts such as cancer detection, where missed diagnoses can delay treatment. 
\[
\text{Recall} = \frac{TP}{TP + FN}.
\]
High recall is essential in clinical contexts such as early cancer detection, where missing positive cases (false negatives) could delay treatment and lead to adverse outcomes.  

Because precision and recall often trade off against each other, we additionally report the F1-score, which summarizes their balance into a single value:
\[
\text{F1} = \frac{2 \cdot \text{Precision} \cdot \text{Recall}}
{\text{Precision} + \text{Recall}}.
\]
Clinically, this combined measure is valuable because high precision reduces the risk of over-diagnosis and unnecessary interventions, while high recall ensures that true disease cases are not missed. The F1-score therefore captures a clinically meaningful balance, rewarding models that can simultaneously avoid excessive false positives and minimize false negatives, which is particularly important in diagnostic workflows where both types of errors carry significant consequences.
  
For multi-class settings, we report the weighted F1-score:
\[
\text{F1}_{\text{weighted}} = \frac{1}{N}\sum_{c=1}^{C} n_c \cdot \text{F1}_c,
\]
where $n_c$ is the number of samples in class $c$, $\text{F1}_c$ is the F1-score for class $c$, and $N$ is the total number of samples. This formulation accounts for class imbalance by ensuring that both majority and minority classes contribute proportionally to the overall evaluation.  

For multi-class classification, we extend the F1-score by computing a weighted average across classes:
\[
\text{F1}_{\text{weighted}} = \frac{1}{N} \sum_{c=1}^{C} n_c \cdot \text{F1}_c,
\]
where $n_c$ is the number of samples in class $c$, $\text{F1}_c$ is the per-class F1-score, $C$ is the total number of classes, and $N$ is the overall sample size. This formulation accounts for class imbalance by assigning greater weight to more prevalent classes while still incorporating performance on minority classes. Clinically, the weighted F1-score ensures that rare but critical categories (e.g., high-grade tumors or rare genetic subtypes) are not overshadowed by abundant negative or benign cases, providing a more faithful reflection of real-world diagnostic performance across heterogeneous patient populations.

\textbf{Advanced robustness metrics.}  
To provide a more comprehensive view of classification performance, particularly under class imbalance and variable decision thresholds, we additionally report the Area Under the ROC Curve (AUC), the Matthews Correlation Coefficient (MCC), and Cohen’s Kappa coefficient.  

The AUC summarizes the Receiver Operating Characteristic curve, which plots the true positive rate (TPR $=TP/(TP+FN)$) against the false positive rate (FPR $=FP/(FP+TN)$) across all possible thresholds. AUC values range from 0.5 (no discriminative ability) to 1.0 (perfect separation). Unlike accuracy or F1, AUC captures global discriminative power independent of a single threshold, making it particularly informative for screening tasks where cut-offs may differ across clinical settings.  

The MCC provides a single balanced statistic that incorporates all four outcomes ($TP$, $TN$, $FP$, $FN$):
\[
\text{MCC} = \frac{TP \cdot TN - FP \cdot FN}
{\sqrt{(TP+FP)(TP+FN)(TN+FP)(TN+FN)}}.
\]
Ranging from $-1$ (inverse prediction) to $+1$ (perfect prediction), MCC is robust to class imbalance and is therefore well-suited for rare disease detection, where conventional accuracy may be inflated by negative cases.  

Finally, Cohen’s Kappa measures agreement between predicted and reference labels while correcting for chance:
\[
\kappa = \frac{p_o - p_e}{1 - p_e},
\]
where $p_o$ is the observed agreement and $p_e$ the expected agreement by chance. With values from $-1$ to $+1$, Kappa offers an interpretable analogue to inter-observer agreement among physicians, with $\kappa > 0.8$ typically regarded as strong concordance in clinical studies.  

\textbf{Summary.}  
Together, these task-level, regression, and classification metrics provide a multifaceted evaluation of TissueLab. While accuracy and error-based measures reflect overall correctness and magnitude of deviation, precision-recall-F1 quantify diagnostic trade-offs, correlation captures consistency in relative ordering, and advanced metrics such as AUC, MCC, and Kappa ensure robustness under imbalance, threshold variability, and human-AI agreement. This spectrum of metrics thus ensures that model performance is not only statistically valid but also clinically meaningful.

\section*{Author Contributions}
\textbf{Conceptualization:} Songhao Li,  Zhi Huang. \textbf{Methodology:} Songhao Li, Jonathan Xu, Tiancheng Bao, Yuxuan Liu, Yuchen Liu, Yihang Liu, Lilin Wang, Wenhui Lei, Sheng Wang, Yinuo Xu, Yan Cui, Jialu Yao, Shunsuke Koga, Zhi Huang. \textbf{Experiments:} Songhao Li, Tiancheng Bao, Wenhui Lei, Yinuo Xu, Zhi Huang. \textbf{Data curation and acquisition:} Songhao Li, Jonathan Xu, Tiancheng Bao, Wenhui Lei, Shunsuke Koga, Zhi Huang. \textbf{Expert Pathology Annotation and Feedback:} Shunsuke Koga. \textbf{Manuscript writing:} Songhao Li, Zhi Huang. \textbf{Supervision:} Zhi Huang. All authors reviewed and approved the final manuscript.

\section*{Acknowledgements}
We would like to thank other software engineers who have contributed to this software, including but not limited to, Steven Su, Qintian Huang. 

\section*{Data Availability}
Whole-slide images and expert annotations from the LNCO2 dataset are available via the AIDA Data Hub at \url{https://datahub.aida.scilifelab.se/10.23698/aida/lnco2} upon application and approval. Public radiology datasets are accessible from their respective repositories: UNIFESP Chest CT Fatty Liver Competition at \url{https://www.kaggle.com/competitions/unifesp-fatty-liver}, NIH Chest X-rays at \url{https://www.kaggle.com/datasets/nih-chest-xrays/data/data}, Sunnybrook Cardiac Data (SCD) at \url{https://www.cardiacatlas.org/studies/sunnybrook-cardiac-data/}, and PhysioNet Intracranial Hemorrhage CT dataset at \url{https://physionet.org/content/ct-ich/1.3.1/}. VisiumHD datasets of prostate, colon, and kidney tissues, including histology slides and matched spatial transcriptomics data, are publicly available from 10x Genomics at \url{https://www.10xgenomics.com/spatial-transcriptomics}.

\section*{Code Availability}
TissueLab source code is publicly available at \href{https://www.tissuelab.org?from=arxiv}{\textcolor{tissuelabblue}{\mbox{tissuelab.org}}}, with installers for Windows, macOS, and Linux.

\section*{Conflict of Interests}
None declared.

\clearpage
\printbibliography
\clearpage
\appendix

\begin{figure}[hbtp]
    \centering
    \includegraphics[width = \textwidth]{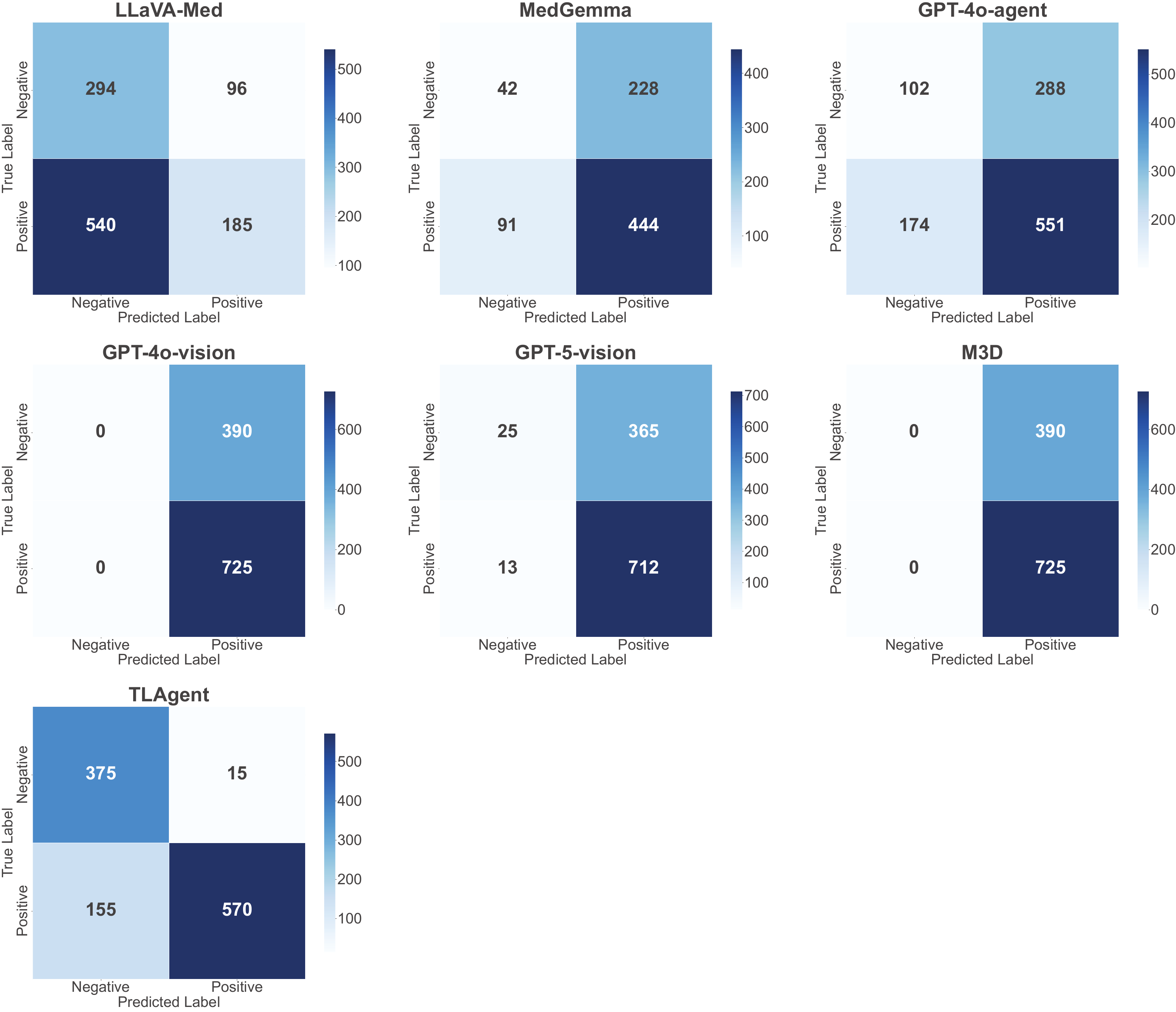}
    \caption{\small \textbf{Confusion matrices for fatty liver diagnosis across models.} 
    Shown are confusion matrices for LLaVA-Med, MedGemma, GPT-4o agent, GPT-4o vision, GPT-5 vision, M3D, and the TissueLab agentic AI system (TLAgent). While some baseline vision-language models appear to achieve reasonable overall metrics, their confusion matrices reveal systematic misclassifications and failure to identify the correct categories. In contrast, TLAgent demonstrates consistent and clinically reliable predictions across classes, underscoring the importance of evaluating performance not only by aggregate scores but also by class-wise diagnostic fidelity.}
    \label{supplementary:3}
\end{figure}

\begin{figure}[hbtp]
    \centering
    \includegraphics[width = \textwidth]{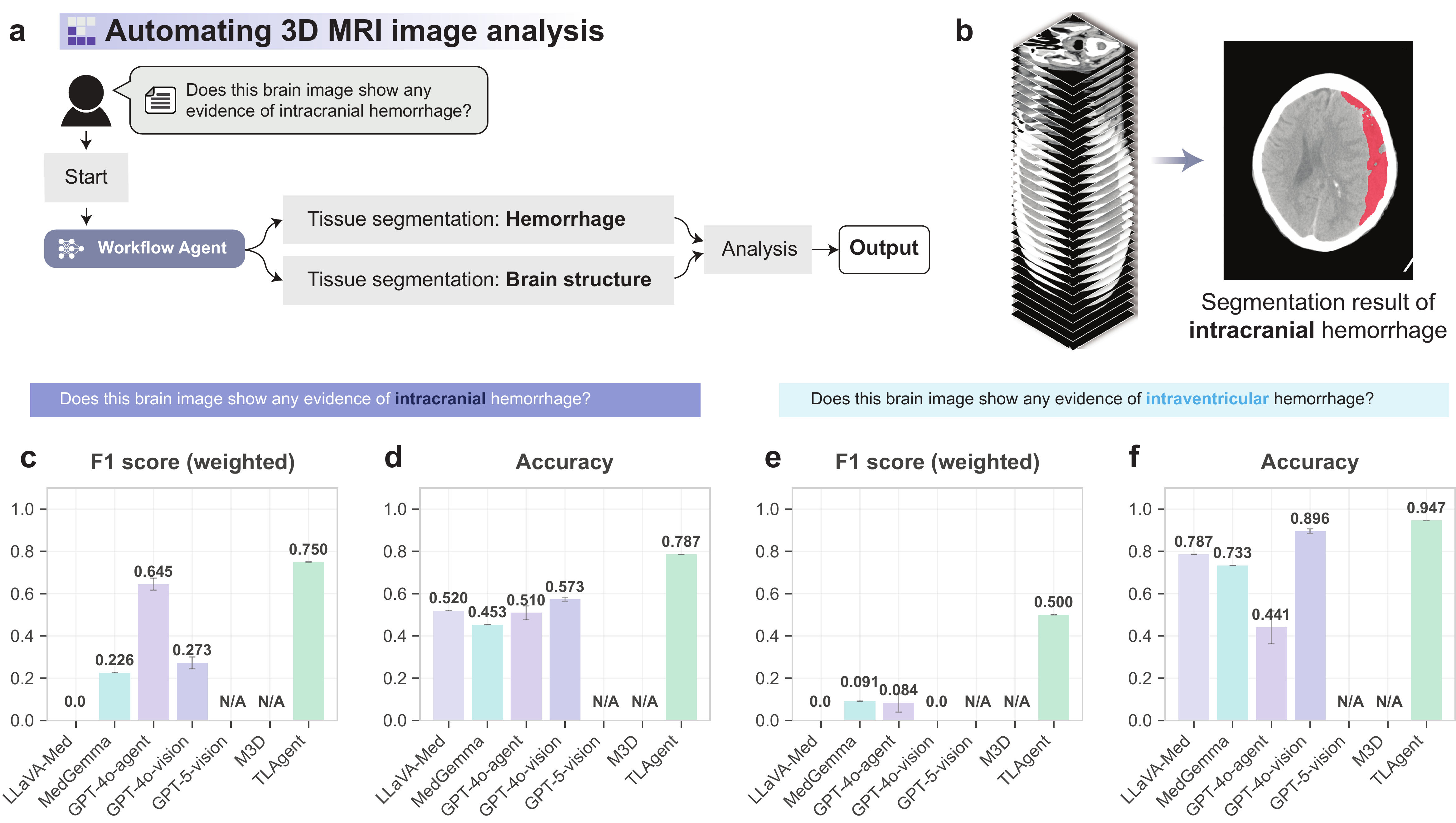}
    \caption{\small \textbf{Automating 3D MRI image analysis for intracranial hemorrhage.} 
    \textbf{a}, Workflow generated by the agent using segmentation of hemorrhage and brain structures as the basis for diagnosis, followed by code-based quantitative analysis. 
    \textbf{b}, Visualization of segmentation results: the left panel shows the full stack of slices, while the right panel displays a representative slice with hemorrhage regions highlighted in red. 
    \textbf{c}, F1-score comparison across models for the primary diagnostic question: \textit{``Does this brain image show any evidence of intracranial hemorrhage?''} 
    \textbf{d}, Accuracy comparison for the same primary diagnostic task. 
    \textbf{e}, F1-score comparison for the fine-grained sub-question: \textit{``Does this brain image show any evidence of intraventricular hemorrhage?''} 
    \textbf{f}, Accuracy comparison for the sub-question task. 
    Collectively, these panels show how the TissueLab agentic AI system automates end-to-end 3D MRI analysis-from workflow construction and visualization to quantitative benchmarking-while delivering more consistent and clinically reliable hemorrhage detection than baseline models.}
    \label{supplementary:4}
\end{figure}
\end{document}